\documentclass[letterpaper]{article} 
\usepackage{aaai23}  
\usepackage{times}  
\usepackage{helvet}  
\usepackage{courier}  
\usepackage[hyphens]{url}  
\usepackage{graphicx} 
\usepackage{natbib}  
\usepackage{caption} 
\usepackage{algorithm}
\usepackage{algorithmic}
\usepackage{multirow}
\usepackage{microtype}
\usepackage{graphicx}
\usepackage{subfigure}
\usepackage{booktabs}

\usepackage{amsmath,bm}
\usepackage{amssymb}
\usepackage{mathtools}
\usepackage{amsthm}

\newtheorem{theorem}{Theorem}
\newtheorem{lemma}{Lemma}
\newtheorem{definition}{Definition}

\usepackage{newfloat}
\usepackage{listings}
\DeclareCaptionStyle{ruled}{labelfont=normalfont,labelsep=colon,strut=off} 
\floatstyle{ruled}
\newfloat{listing}{tb}{lst}{}
\floatname{listing}{Listing}
\title{Adversarial Weight Perturbation Improves Generalization \\in Graph Neural Networks}
\author {
   Yihan Wu,\textsuperscript{\rm 1} 
    Aleksandar Bojchevski,\textsuperscript{\rm 2} 
    Heng Huang\textsuperscript{\rm 1}
}
\affiliations {
    \textsuperscript{\rm 1} Electrical and Computer Engineering, University of Pittsburgh, PA, USA\\
    \textsuperscript{\rm 2} CISPA Helmholtz Center for Information Security\\
    yiw154@pitt.edu, bojchevski@cispa.de, henghuanghh@gmail.com
}

\usepackage{bibentry}

\begin{document}

\maketitle

\begin{abstract}
A lot of theoretical and empirical evidence shows that the flatter local minima tend to improve generalization. Adversarial Weight Perturbation (AWP) is an emerging technique to efficiently and effectively find such minima. In AWP we minimize the loss w.r.t. a bounded worst-case perturbation of the model parameters thereby favoring local minima with a small loss in a neighborhood around them.
The benefits of AWP, and more generally the connections between flatness and generalization, have been extensively studied for \emph{i.i.d.} data such as images. In this paper, we extensively study this phenomenon for graph data. Along the way, we first derive a generalization bound for \emph{non-i.i.d.} node classification tasks. Then we identify a vanishing-gradient issue with all existing formulations of AWP and we propose a new Weighted Truncated AWP (WT-AWP) to alleviate this issue. We show that regularizing graph neural networks with WT-AWP consistently improves both natural and robust generalization across many different graph learning tasks and models. Our code is available at \texttt{https://github.com/YihanWu95/WT-AWP}.
\end{abstract}

\section{Introduction}
Simply minimizing the standard cross-entropy loss for highly non-convex and non-linear models such as (deep) neural networks is not guaranteed to obtain solutions that generalize well, especially for today's overparamatrized networks. The key underlying issue is that these models have many different local minima which can have wildly different generalization properties despite having nearly the same performance on training and validation data. Naturally, there is a rich literature that studies the properties of well-behaving local minima, as well as the design choices that improve our chances of finding them \cite{stutz2021relating}. The notion of flatness which measure how quickly the loss changes in a neighbourhood around a given local minimum has been empirically shown to correlate with generalization among a variety of different measures \cite{jiang2019fantastic}. In addition, generalization bounds based on the PAC-Bayes framework \cite{mcallester1999pac, foret2020sharpness} provide theoretical insights that corroborate the mounting empirical data.       
Since the evidence implies that flatter minima tend to generalize better, the obvious question is how to efficiently find them.\looseness=-1

Not only do flat minima improve generalization to unseen test data, i.e. the clean accuracy \cite{foret2020sharpness, zheng2021regularizing, xu2022detached, kwon2021asam, xu2022closing}, but they also improve generalization to adversarial examples, i.e. the robust accuracy \cite{wu2020adversarial, stutz2021relating,wu2022towards}. Improving adversarial robustness is important, especially for models deployed in safety-critical domains,
since most standard (undefended) models are vulnerable to adversarial attacks. Attackers can easily craft deliberate and unnoticeable input perturbations that change the prediction of the classifier \cite{sun2018adversarial}.

Flat minima show higher resistance to adversarially perturbed inputs while maintaining good clean accuracy \cite{stutz2021relating}.
Among the variety of techniques for finding flat minima
Adversarial Weight Perturbation (AWP) \cite{wu2020adversarial},
and the closely-related (adaptive) sharpness-aware minimization \cite{foret2020sharpness, kwon2021asam} and adversarial model perturbation \cite{zheng2021regularizing}, seems to be quite effective in practice. The key idea is to minimize the loss w.r.t. a bounded worst-case perturbation of the model parameters, i.e. minimize a local notion of sharpness. The benefits of this approach, and more generally the correlation between flatness and (clean/robust) generalization, have been extensively studied for i.i.d.  data such as images. In this paper we study this phenomenon for graph data. 
Concretely, we analyze and improve the generalization of Graph Neural Networks (GNNs) which are a fundamental building block (in addition to CNNs and RNNs).\looseness=-1

Blindly applying existing weight perturbation techniques to GNNs is unfortunately not effective in practice due to a vanishing-gradient issue. Intuitively, the adversarially perturbed weights tend to have a higher norm which in turn leads to a saturation in the last layer where that logits for one class are on a significantly larger scale compared to the rest.
Even though this limitation plagues all formulations of AWP, for both GNNs and other models (e.g. ResNets), it has gone unnoticed so far. To address it we propose Weighted Truncated Adversarial Weight Perturbation (WT-AWP) where rather than directly minimizing the (robust) AWP loss we use it as a regularizer in addition to the standard cross-entropy loss.
Moreover, we propose to abstain from perturbation in the last layer(s) of the network for a more fine-grained control of the training dynamics. These two modifications are simple, but necessary and effective. With our resulting formulation the models can obtain useful gradient signals for training even when the perturbed weights have a high norm, mitigating the gradient-vanishing issue. Furthermore, we theoretically study the AWP learning objective and show its invariance for local extrema. We can summarize our contributions as follows:
\begin{itemize}
    \item We provide a theoretical analysis of AWP on non-i.i.d. tasks and identify a vanishing-gradient issue that plagues all previous AWP variants. Based on this analysis we propose Weighted Truncated Adversarial Weight Perturbation (WT-AWP) that mitigates this issue.
    \item We study 
    the connections between flatness and generalization for Graph Neural Networks. We show that GNNs trained with our WT-AWP formulation have simultaneously improved natural and robust generalization. The improvement is statistically significant and consistent across tasks (node-level and graph-level classification) and across models (standard and robustness-aware GNNs), at a negligible computational cost.
\end{itemize}

\section{Background and Related Work}\label{sec:relatedw}
\textbf{Adversarial Weight Perturbation for Images.} 
AWP is motivated by the connection between the flatness of the loss landscape and model generalization. Given a learning objective $L(\cdot)$ and an image classification model with parameters ${\bm{\theta}}$, the generalization gap \cite{wu2020adversarial}, also named the sharpness term  \cite{foret2020sharpness}, which measures the worst-case flatness of the loss landscape, is defined by $[\max_{||{\bm{\delta}}||\leq\rho}L({\bm{\theta}}+{\bm{\delta}}) -L({\bm{\theta}})]$. This gap is known to control a PAC-Bayes generalization bound \cite{neyshabur2017exploring}, with a smaller gap implying better generalization.
The AWP objective simultaneously minimizes the loss function and the generalization gap via
$\min_{{\bm{\theta}}}[L({\bm{\theta}})+(\max_{||{\bm{\delta}}||\leq\rho}L({\bm{\theta}}+{\bm{\delta}}) -L({\bm{\theta}}))] = \min_{{\bm{\theta}}}\max_{||{\bm{\delta}}||\leq\rho}L({\bm{\theta}}+{\bm{\delta}})$.
Providing further theoretical justification for the effectiveness of the AWP, \cite{zheng2021regularizing} prove that this objective favors solutions corresponding to flatter local minima assuming that the loss surface can be approximated as an inverted Gaussian surface. Relatedly, they show that AWP penalizes the gradient-norm. 

In some cases we can rescale the weights to achieve arbitrarily sharp minima that also generalize well \cite{dinh2017sharp}. We can mitigate this issue using a scale-invariant definition of sharpness \cite{kwon2021asam}. Since in our experiments such adaptive sharpness was not beneficial we present the non-adaptive case for simplicity but all results can be trivially extended.
\citet{keskar2016large} show that large-batch training may reach sharp minima, however, this does not affect GNNs since they tend to use a small batch size. \looseness=-1

\noindent\textbf{GNNs, Graph attacks, and Graph defenses.}
Graph Neural Networks (GNNs) are emerging as a fundamental building block. They have achieved spectacular results on a variety of graph learning tasks across many high-impact domains (see survey \cite{wu2020comprehensive}).
Despite their success, it has been demonstrated that GNNs suffer from evasion attacks at test time \cite{zugner2018adversarial} and poisoning attacks at training time \cite{zugner2019adversarial}.
Meanwhile, a series of methods have been developed to improve their robustness. For example, GCNJaccard \cite{wu2019adversarial} drops dissimilar edges in the graph, as it found that attackers tend to add edges between nodes with different features. GCNSVD \cite{entezari2020all} 
replaces the adjacency matrix with its low-rank approximation motivated by the observation that mostly the high frequency spectrum of the graph is affected by the adversarial perturbations. We also have provable defenses that provide robustness certificates \cite{bojchevski2020efficient}. 
Both heuristic defenses (e.g. GCNJaccard and GCNSVD) and certificates are improved with our WT-AWP.
For an overview of attacks and defenses see \citet{sun2018adversarial}.\looseness=-1
\section{Adversarial Weight Perturbation on GNNs}\label{sec:gnnawp}
To simplify the exposition we focus on the semi-supervised node classification task. Nonetheless, in Sec.~\ref{sec:graph_classification} we show that AWP also improves graph-level classification.
Let $G = (\bm{A},\bm{X})$ be a given (attributed) graph where $\bm{A}$ is the adjacency matrix and $\bm{X}$ contains the node attributes. 
Let $\mathcal{V}$ be the set of all nodes. In semi-supervised node classification problem we have access to the entire graph, the features and neighbors for all nodes $\mathcal{V}$, but we only have labels for a (small) subset of $\mathcal{V}$ (usually 10\%). 
Normally we optimize $\min_{{\bm{\theta}}} L_\text{train}({\bm{\theta}};\bm{A},\bm{X})$,
where $L_\text{train}=\sum_{v\in\mathcal{V_{\text{train}}}} l(f_{\bm{\theta}}(\bm{A},\bm{X}), y_v)$, $f$ is a GNN parametrized by weights ${\bm{\theta}}=(\bm{\theta}_1,...,\bm{\theta}_k)$, $y_v$ is the ground-truth label for node $v$, and $l$ is some loss function (e.g. cross-entropy) applied to each node in the training set $\mathcal{V}_\text{train} \subset \mathcal{V}$.

In AWP we first find the worst-case weight perturbation ${\bm{\delta}}^*({\bm{\theta}})$ that maximizes the loss. Then we minimize the loss with the perturbed weights.
The worst-case perturbation for a given ${\bm{\theta}}$ is defined as
\begin{equation}\label{eq:wp}
 {\bm{\delta}}^*({\bm{\theta}}) := \textrm{arg}\max_{
 ||\bm{\delta}||_2\leq\rho
 }L_\text{train}({\bm{\theta}}+{\bm{\delta}};\bm{A},\bm{X})
 \end{equation}
where
$\rho$ 
is the strength of perturbation. The AWP learning objective is then $\min_{{\bm{\theta}}}\max_{||\bm{\delta}||\leq\rho}L_\text{train}({\bm{\theta}}+{\bm{\delta}};\bm{A},\bm{X})$, or
\begin{equation}\label{eq:awp}
\begin{aligned}
\min_{{\bm{\theta}}}L_\text{train}({\bm{\theta}}+{\bm{\delta}}^*({\bm{\theta}});\bm{A},\bm{X}).
\end{aligned}
\end{equation}
Since the PAC-Bayes bound proposed by \citet{mcallester1999pac} only holds for i.i.d. data and semi-supervised node classification is a non-i.i.d. task, the analyses in \citet{wu2020adversarial} and \citet{foret2020sharpness} cannot be directly extended to node classification. Thus, we derive a new generalization bound for node classification (with GNNs) based on a recent sub-group generalization bound \cite{ma2021subgroup}.

\begin{theorem}[Generalization bound of AWP loss] Let $L_\textup{all}({\bm{\theta}};\bm{A},\bm{X})$ be the loss on all nodes, for any set of training nodes $\mathcal{V}_\textup{train}$ from $\mathcal{V}$, $\forall m\geq\sqrt{d}$, with probability at least $1-\delta$, we have
\begin{equation}\label{eq:pacbayes1}
\begin{aligned}
    &L_{\textup{all}}(\bm{\theta};\bm{A},\bm{X})\leq\max_{||\bm{\delta}||_2\leq \rho}[L_{\textup{train}}(\bm{\theta}+\bm{\delta};\bm{A},\bm{X})]\\
    &+(\frac{m^2}{d}e^{1-\frac{m^2}{d}})^{d/2}
+\frac{1}{2\sqrt{N_0}}\left[1+d\log(1+\frac{m^2||\bm{\theta}||_2^2}{d\rho^2})\right]\\
&+\frac{1}{\sqrt{N_0}}\left(\ln\frac{3}{\delta}+\frac{1}{4}+\Theta(K \cdot \epsilon_\textup{all})\right).
\end{aligned}
\end{equation}
where $d$ is the number of parameters in the GNN, $K$ is the number of groundtruth labels, $\epsilon_\textup{all}$ is a fixed constant w.r.t. $\mathcal{V}$, $N_0$ is the volume of $\mathcal{V}_\textup{train}$, and $\rho$ is the perturbation strength on the weights.
\end{theorem}

The details of the proof are in Sec. \ref{sec:genebound}.
We can rewrite Eq.~\ref{eq:pacbayes1} into the following simplified version
\begin{equation}\label{eq:pacbayes}
\begin{aligned}
\small
     L_\textup{all}({\bm{\theta}};\bm{A},\bm{X})\leq \max_{||{\bm{\delta}}||_2\leq \rho}L_\textup{train}({\bm{\theta}}+{\bm{\delta}};\bm{A},\bm{X})
     +h(||{\bm{\theta}}||_2^2/\rho^2)
\end{aligned}
\end{equation}
where $h(\cdot)$ is a monotonously increasing function depending on the perturbation strength $\rho(\bm{\theta})$.

This bound justifies the use of AWP since the perturbed loss on training nodes bounds the standard loss on \emph{all} nodes. Moreover, as $h(||{\bm{\theta}}||_2^2/\rho^2)$ is monotonically decreasing with $\rho$, increasing the perturbation strength $\rho$ can make the bound in Eq. \ref{eq:pacbayes} sharper, i.e. the resulting AWP objective should lead to better generalization.
In practice we perturb the weights $\bm{\theta_i}$ of each layer separately, and this bound still holds if we set $\rho =
\sum_{i=1}^{k}\rho(\bm{\theta_i})$ where $\rho(\bm{\theta_i})$ is the perturbation strength for layer $i$.
We derived a similar result for graph-level tasks in Sec. \ref{sec:gen_gap_graph_level}.

Since finding the optimal perturbation (Eq. \ref{eq:wp}) is intractable, we approximate it with a one-step projected gradient descent as in previous work \cite{wu2020adversarial,foret2020sharpness,zheng2021regularizing},\looseness=-1 
 \begin{equation}\label{eq:est}
\hat{{\bm{\delta}}}^*({\bm{\theta}}) := \Pi_{B(\rho({\bm{\theta}}))}( \nabla_{\bm{\theta}} L_\text{train}({\bm{\theta}};\bm{A},\bm{X})),
 \end{equation}
 where $B(\rho({\bm{\theta}}))$ is an $l_2$ ball with radius $\rho({\bm{\theta}})$ and $\Pi_{B(\rho({\bm{\theta}}))}(\cdot)$ is a projection operation, which projects the perturbation back to the surface of $B(\rho({\bm{\theta}}))$ when the perturbation is out of the ball. The maximum perturbation norm $\rho({\bm{\theta}})$ could either be a constant \cite{foret2020sharpness,zheng2021regularizing} or layer dependent \cite{wu2020adversarial}. We specify a layer-dependent norm constraint $\rho({\bm{\theta_i}}) := \rho||{\bm{\theta_i}}||_2$ because the scales of different layers in a neural network can vary greatly.
 With the approximation $\hat{\bm{\delta}}^*({\bm{\theta}})$, the definition of the final AWP learning objective is given by
  \begin{equation}\label{eq:awpest}
  \begin{aligned}
 L_\text{awp}({\bm{\theta}}):=
 L_\text{train}({\bm{\theta}}+\Pi_{B(\rho({\bm{\theta}}))}( \nabla_{\bm{\theta}} L_\text{train}({\bm{\theta}};\bm{A},\bm{X});\bm{A},\bm{X}),
  \end{aligned}
 \end{equation}
 If $L_\text{train}({\bm{\theta}};\bm{A},\bm{X})$ is smooth enough, $\nabla_{\bm{\theta}} L_\text{train}({\bm{\theta}};\bm{A},\bm{X})=0$ when ${\bm{\theta}}^{*}$ is a local minimum. In this case  $L_\text{awp}({\bm{\theta}})=L_\text{train}({\bm{\theta}};\bm{A},\bm{X})$. A natural question is whether ${\bm{\theta}}^{*}$ will also be the minimum of $L_\text{awp}({\bm{\theta}})$?
 We show that $L_\text{awp}({\bm{\theta}})$ keeps the local minimum of $L_\text{train}({\bm{\theta}};\bm{A},\bm{X})$ unchanged.
\begin{theorem}
\label{th:invariant}
(Invariant of local minimum) With the AWP learning objective in Eq. \ref{eq:awpest}, and for continuous  $L_\textup{train}({\bm{\theta}};\bm{A},\bm{X})$, $\nabla_{{\bm{\theta}}}L_\textup{train}({\bm{\theta}};\bm{A},\bm{X})$, ${\Delta}_{{\bm{\theta}}}L_\textup{train}({\bm{\theta}};\bm{A},\bm{X})$, if ${\bm{\theta}}^*$ is a local minimum of $L_\textup{train}({\bm{\theta}};\bm{A},\bm{X})$ and the Hessian matrix ${\Delta}_{{\bm{\theta}}}L_\textup{train}({\bm{\theta}};\bm{A},\bm{X})|_{{\bm{\theta}}*}$ is positive definite, ${\bm{\theta}}^*$ is also a local minimum of $L_\textup{awp}({\bm{\theta}})$.
\end{theorem}

 

The proof is provided in Sec. \ref{sec:proofs}. 
The exact gradient of this new objective is 
 \begin{align}\label{eq:grad_exact}
 &\nabla_{\bm{\theta}} L_\text{train}({\bm{\theta}}+\hat{{\bm{\delta}}}^*({\bm{\theta}});\bm{A},\bm{X})=\nabla_{\bm{\theta}} L_\text{train}({\bm{\theta}};\bm{A},\bm{X})|_{{\bm{\theta}}+\hat{{\bm{\delta}}}^*({\bm{\theta}})} \nonumber \\
 &+\nabla_{\bm{\theta}}\hat{{\bm{\delta}}}^*({\bm{\theta}})\nabla_{\bm{\theta}} L_\text{train}({\bm{\theta}};\bm{A},\bm{X})|_{{\bm{\theta}}+\hat{{\bm{\delta}}}^*({\bm{\theta}})}
 \end{align}
 Since $\nabla_{\bm{\theta}}\hat{{\bm{\delta}}}^*({\bm{\theta}})$ includes second and higher order derivative of ${\bm{\theta}}$, which are computationally expensive, they are omitted during training, obtaining the following approximate gradient of the AWP loss
 \begin{equation}\label{eq:grad_awp}
 \begin{aligned}
 \nabla_{\bm{\theta}} L_\text{train}({\bm{\theta}};\bm{A},\bm{X})|_{{\bm{\theta}}+\hat{{\bm{\delta}}}^*({\bm{\theta}})}
 \end{aligned}
 \end{equation}
\citet{foret2020sharpness} show the models trained with the exact gradient (Eq. \ref{eq:grad_exact}) have almost the same performance as model trained with the approximate first-order gradient (Eq. \ref{eq:grad_awp}). Besides, we can also show that the norm of the difference between Eq. \ref{eq:grad_exact} and Eq. \ref{eq:grad_awp} is proportional to $\rho$ and the second order derivatives of loss the $L$ w.r.t. the weights $\theta$.

\section{Weighted Truncated AWP}
In this section we discuss the theoretical limitations of existing AWP methods on GCN, and illustrate them empirically on a toy dataset. We also propose two approaches to improve AWP. Our improved AWP works well on both toy data and on real-world GNN benchmarks across many tasks and models. We also show that similar problems also exist for multi-layer perceptrons (see Sec. \ref{sec:vanishmlp}).

\subsection{The Vanishing-gradient Issue of AWP}
\label{sec:vanishing_gradient}

\begin{figure*}[!t]
	\subfigure[Vanilla GCN]{
		\begin{minipage}[t]{.23\linewidth}
			\centering
			\includegraphics[height=2.5cm]{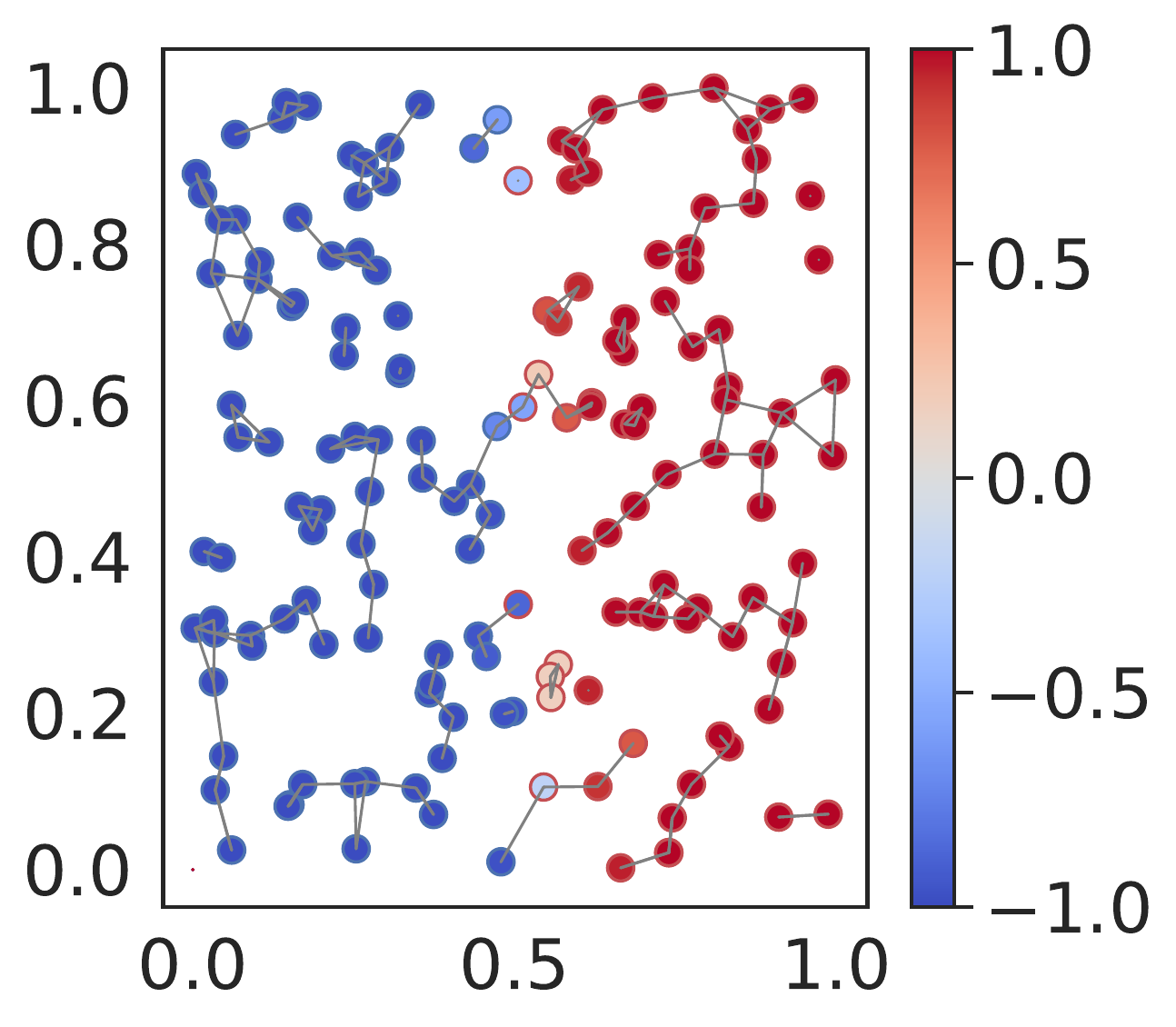}	
			\label{fig:1a}
		\end{minipage}
	}
	\subfigure[AWP $\rho = 0.5$]{
		\begin{minipage}[t]{.23\linewidth}
			\centering
			\includegraphics[height=2.5cm]{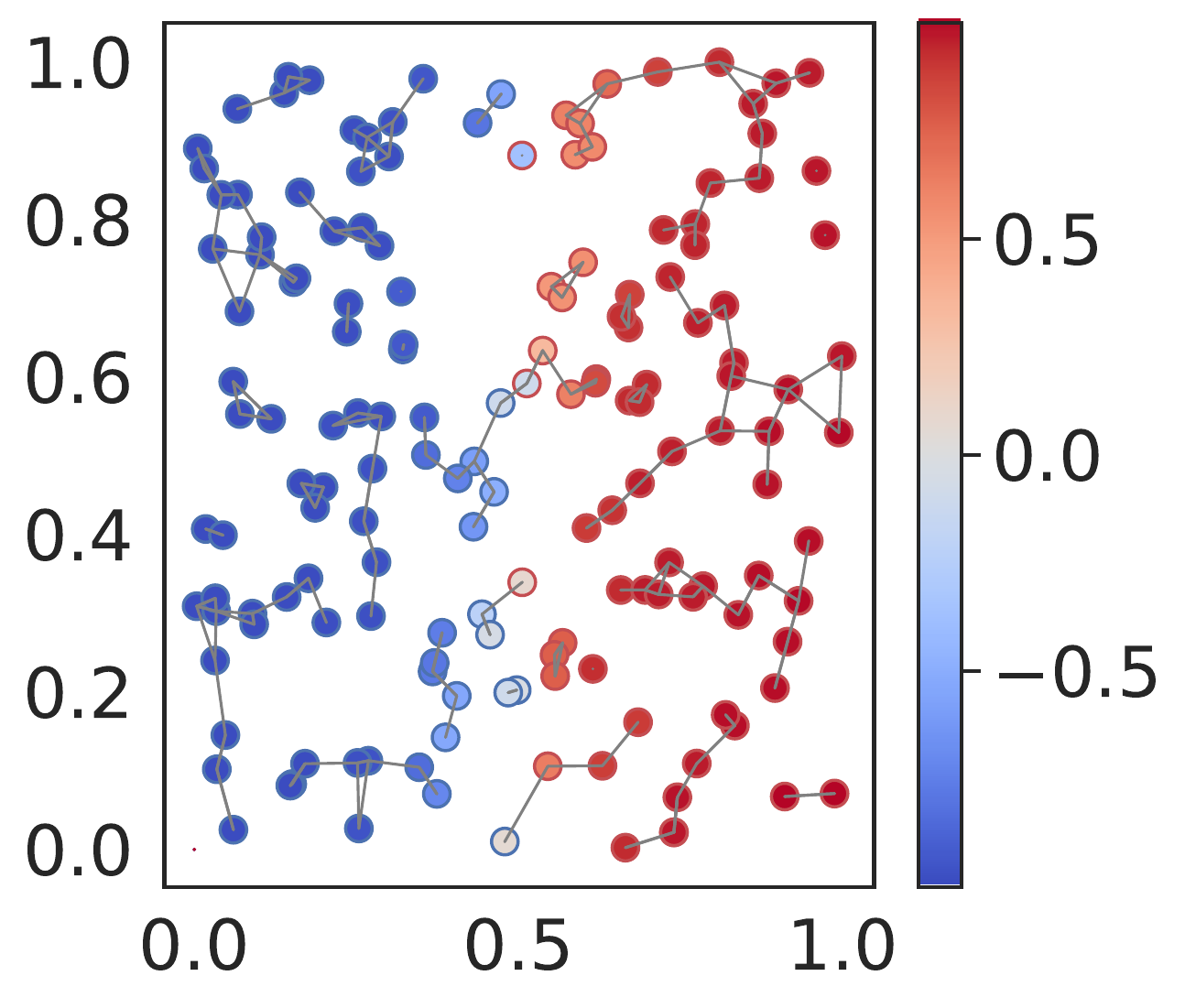}	
			\label{fig:1b}
		\end{minipage}
	}
	\subfigure[AWP $\rho = 1.5$]{
		\begin{minipage}[t]{.23\linewidth}
			\centering
			\includegraphics[height=2.5cm]{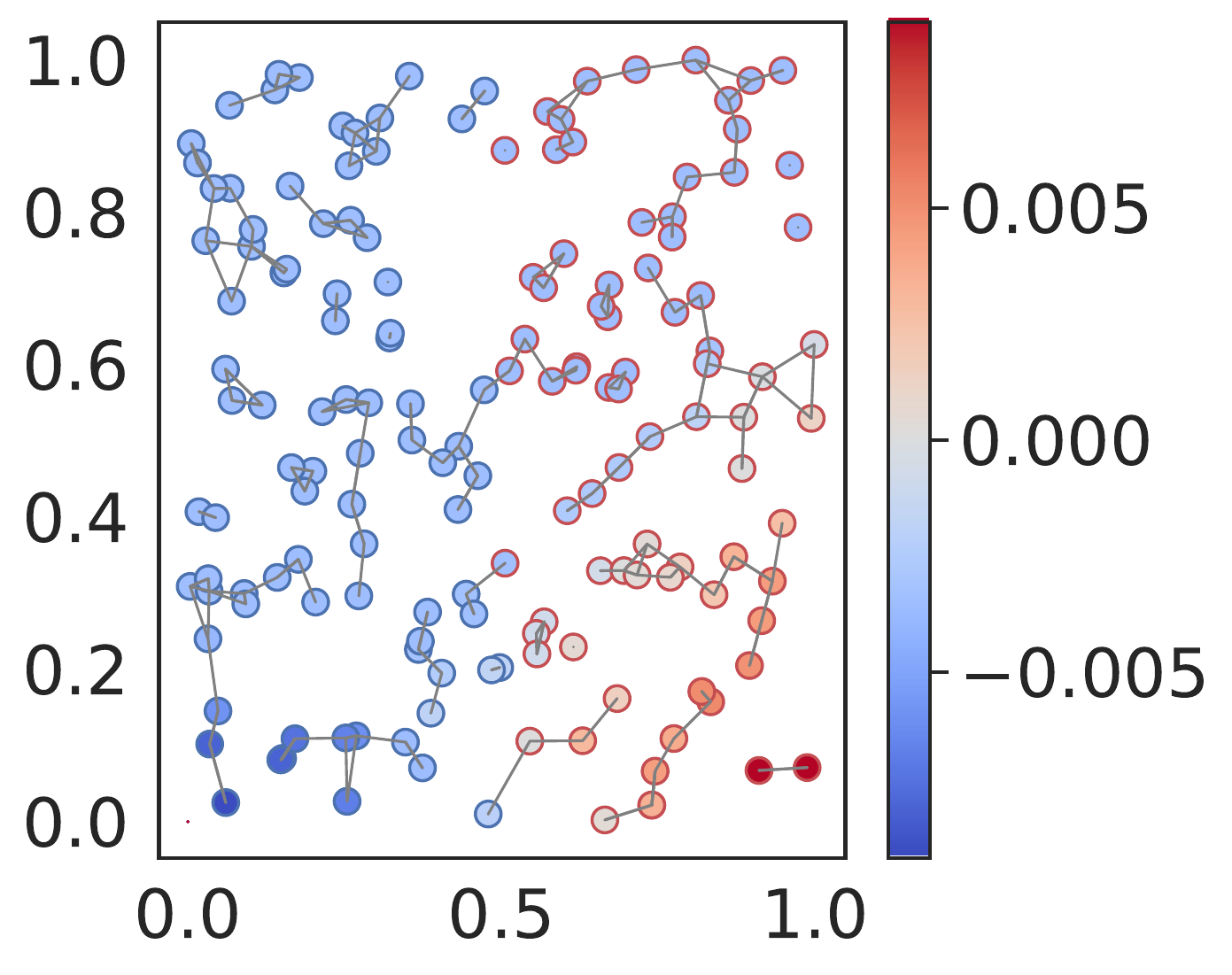}	
			\label{fig:1c}
		\end{minipage}
	}
	\subfigure[AWP $\rho = 2.5$]{
		\begin{minipage}[t]{.23\linewidth}
			\centering
			\includegraphics[height=2.5cm]{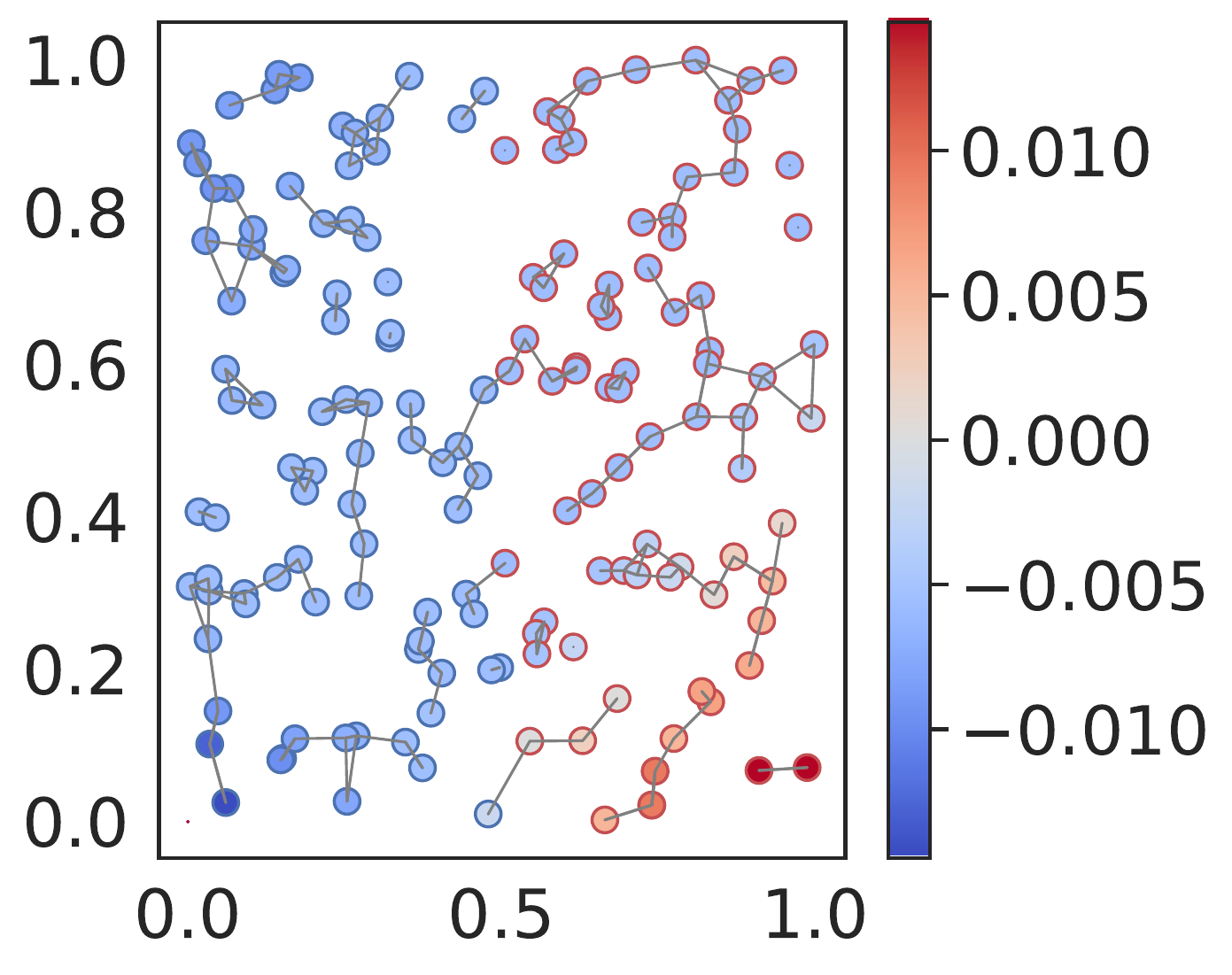}	
			\label{fig:1d}
		\end{minipage}
	}
	\vspace{-10pt}
	\caption{Compare AWP models on a linearly separable dataset with different perturbation strengths $\rho$. The accuracy of models (a) to (d) is 0.97, 0.97, 0.69, and 0.48 respectively. The face color of each node shows its prediction score and the border color shows its ground-truth label. Grey lines connect the node with its nearest neighbours in the graph. For large values of $\rho$ the model is unable to learn.}
	\label{fig:1}
	\vspace{-15pt}
\end{figure*}

Consider a GCN $\hat{y}=\sigma_s(\hat{\bm{A}}(...\sigma(\hat{\bm{A}}\bm{X}\bm{W}_1)...)\bm{W}_n)$ with a softmax activation $\sigma_s$ at the output layer and non-linearity $\sigma$, where $\hat{\bm{A}}$ is the graph Laplacian given by $\hat{\bm{A}}:=\bm{D}^{-1/2}(\bm{A}+\bm{I}_N)\bm{D}^{-1/2}, D_{ii} = \sum_{j}(\bm{A}+\bm{I}_N)_{ij}$. The perturbed model is $\hat{y}=\sigma_s(\hat{\bm{A}}(\dots\sigma(\hat{\bm{A}}\bm{X}(\bm{W}_1+{\bm{\delta}}_1))\dots)(\bm{W}_n+{\bm{\delta}}_n))$. Since the norm of each perturbation ${\bm{\delta}}_i$ could be as large as $\rho||\bm{W}_i||_2$, in the worst case the norm of each layer is $(\rho+1)||\bm{W}_i||_2$, and thus the model will have exploding logit values when $\rho$ is large. If additionaly the logit for one class is significantly larger than the others, the output will approximate a one-hot encoded vector after the softmax. In this case the gradient will be close to $0$ and the weights will not be updated. Although in practice the number of GCN layers is often less than 3, we still observe the vanish gradient issue in both toy datasets and GNN benchmarks.

\begin{figure}
	\centering
	\includegraphics[height=3.5cm]{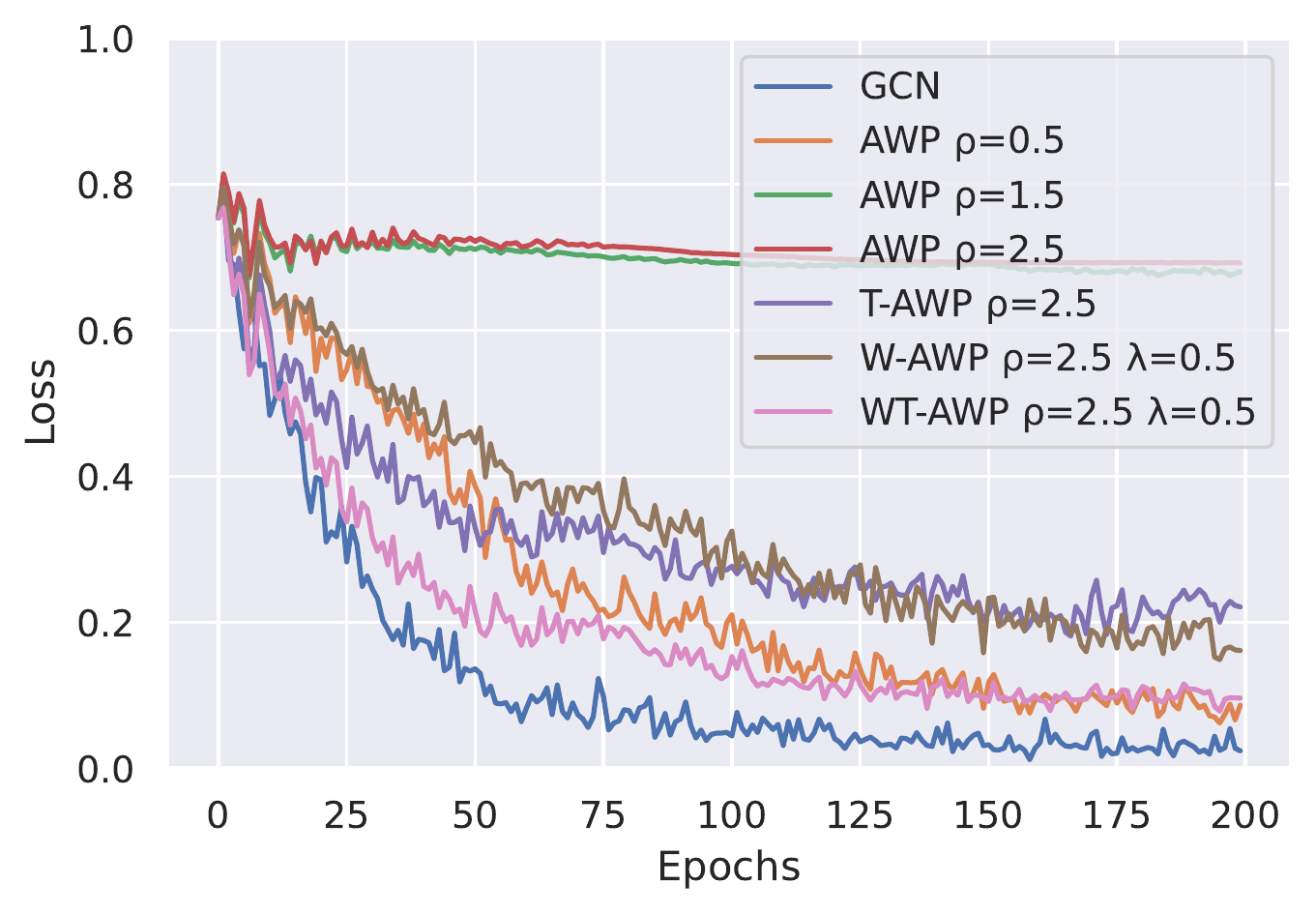}	
	\vspace{-0.3cm}
	\caption{Learning curves for GCN with different losses (and $\rho$).}
	\label{fig:awpknnloss}
	\vspace{-0.6cm}
\end{figure}
To verify our conclusion, we train a 2-layer GCN network with hidden dimension 64, which is a common setting for GCNs, on a linearly separable dataset. The dataset contains 2 classes $\{-1,1\}$ and each class has 100 nodes. We apply $k$-nearest neighbor $(k=3)$ to obtain the adjacency matrix, and use the 2D position of the nodes as the features.
The number of training epochs is 200. We use 10\% nodes for training, 10\% for validating and the rest 80\% for testing.
In Figure \ref{fig:1} we show the trained classifiers for different $\rho$ values. Models with AWP crash quickly as $\rho$ increases from $0.5$ to $2.5$. When $\rho = 0.5$, the classification accuracy is 0.97, which is nearly the same as the vanilla model, but when  $\rho = 2.5$, the classification accuracy is 0.51, which is the same as a random guess. Besides,  when $\rho = 1.5$ and $2.5$, the loss of AWP method is almost constant during training (Figure \ref{fig:awpknnloss}) and the prediction score (Figure \ref{fig:1c} and Figure \ref{fig:1d}) is around $0$. This indicates that the weights are barely updated during training. So with the AWP objective, we cannot select a large $\rho$. Yet, as we discussed in Sec. \ref{sec:gnnawp}, we prefer larger values of $\rho$ since they lead to a tighter bound (Eq. \ref{eq:pacbayes}) and are more like to generalize better. As we shown next, our suggested improvements fix this issue. 

\subsection{Truncated AWP and Weighted AWP} \label{sec:wtawp}
 \textbf{Intuition for WT-AWP.} 
 The vanishing gradient is mainly due to the exploding of the logit values, which is caused by perturbing all layers in the model. Thus, a natural idea is to only apply AWP on certain layers to mitigate the issue. This it the truncated AWP. Another idea is to provide a second source of valid gradients which we do by adding the the vanilla loss $L_\text{train}({\bm{\theta}};\bm{A},\bm{X})$ to the AWP loss. Even when the AWP loss suffers from the vanishing gradient issue, the vanilla loss is not affected.
 



\begin{definition}
(Truncated AWP) We split the model parameters into two parts ${\bm{\theta}} = [{\bm{\theta}}^\textup{(awp)},{\bm{\theta}}^\textup{(normal)}]$, and we only perform AWP on ${\bm{\theta}}^\textup{(awp)}$. The Truncated AWP objective is
\begin{equation}
    \min_{{\bm{\theta}}} L_\textup{train}({\bm{\theta}}+[\hat{\bm{\delta}}^\textup{(awp)*}({\bm{\theta}}^\textup{(awp)}),0];\bm{A},\bm{X}),
    \vspace{-2pt}
\end{equation}
where $\hat{\bm{\delta}}^\textup{(awp)*}({\bm{\theta}}^\textup{(awp)})$ is defined as in Eq. \ref{eq:est}.
\end{definition}

\begin{figure*}[t]
	\subfigure[Vanilla model]{
		\begin{minipage}[t]{.18\linewidth}
			\centering
			\includegraphics[height=2.2cm]{figures/normalknn.pdf}	
			\label{fig:2a}
		\end{minipage}
	}
	\subfigure[AWP]{
		\begin{minipage}[t]{.18\linewidth}
			\centering
			\includegraphics[height=2.2cm]{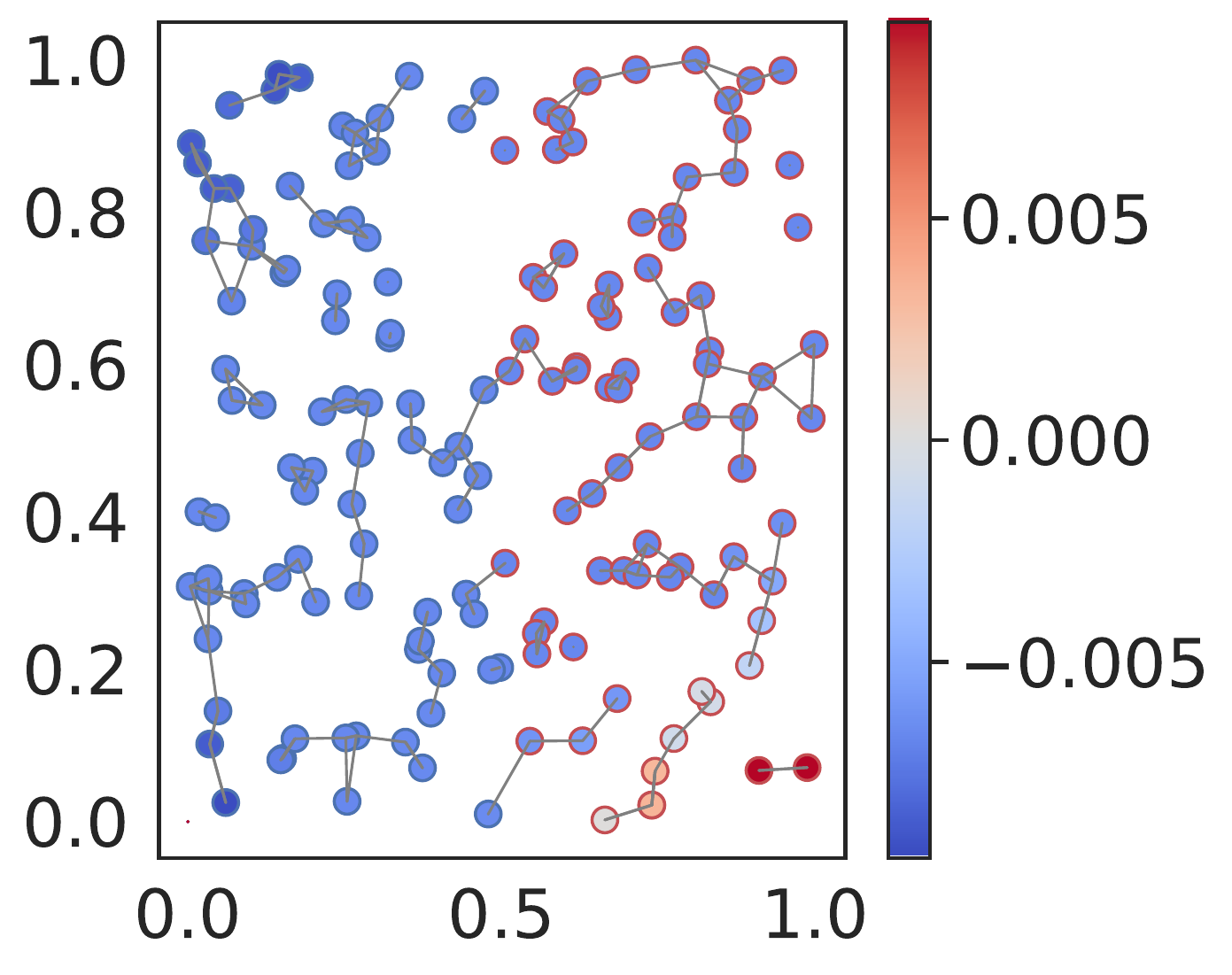}	
			\label{fig:2b}
		\end{minipage}
	}
	\subfigure[T-AWP]{
		\begin{minipage}[t]{.18\linewidth}
			\centering
			\includegraphics[height=2.2cm]{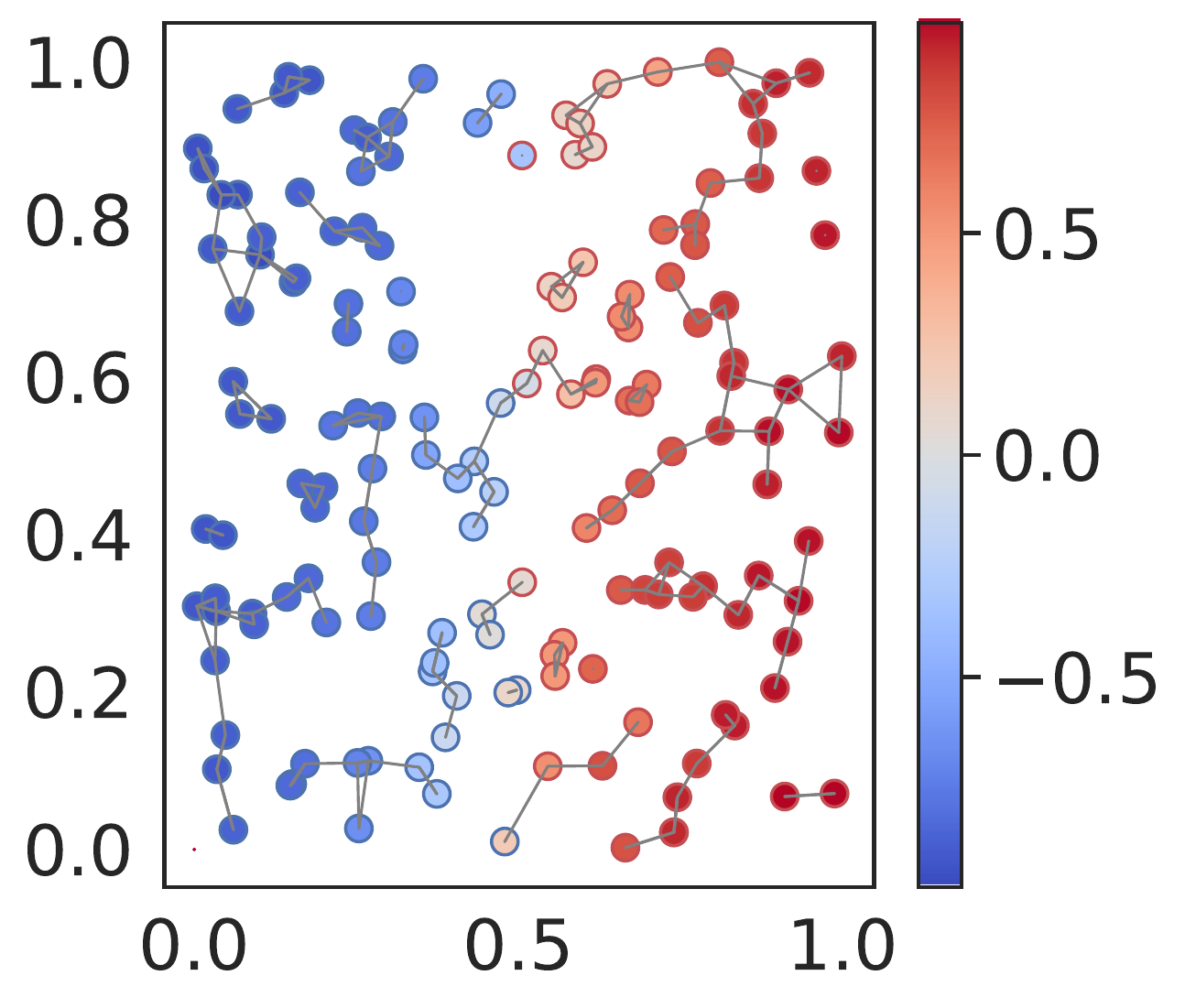}	
			\label{fig:2c}
		\end{minipage}
	}
	\subfigure[W-AWP, $\lambda{=}0.5$]{
		\begin{minipage}[t]{.18\linewidth}
			\centering
			\includegraphics[height=2.2cm]{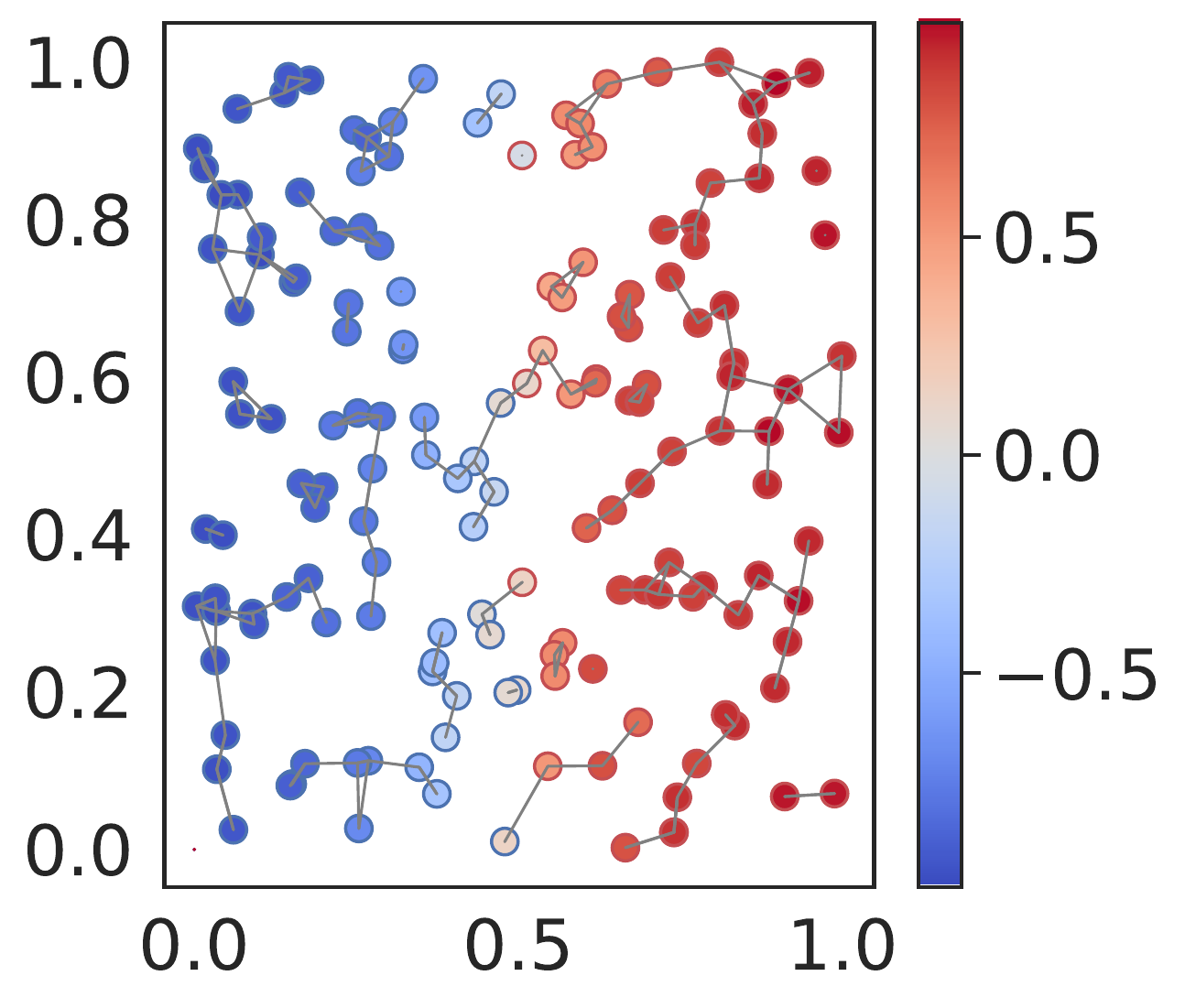}	
			\label{fig:2d}
		\end{minipage}
	}
	\subfigure[WT-AWP, $\lambda{=}0.5$]{
		\begin{minipage}[t]{.18\linewidth}
			\centering
			\includegraphics[height=2.2cm]{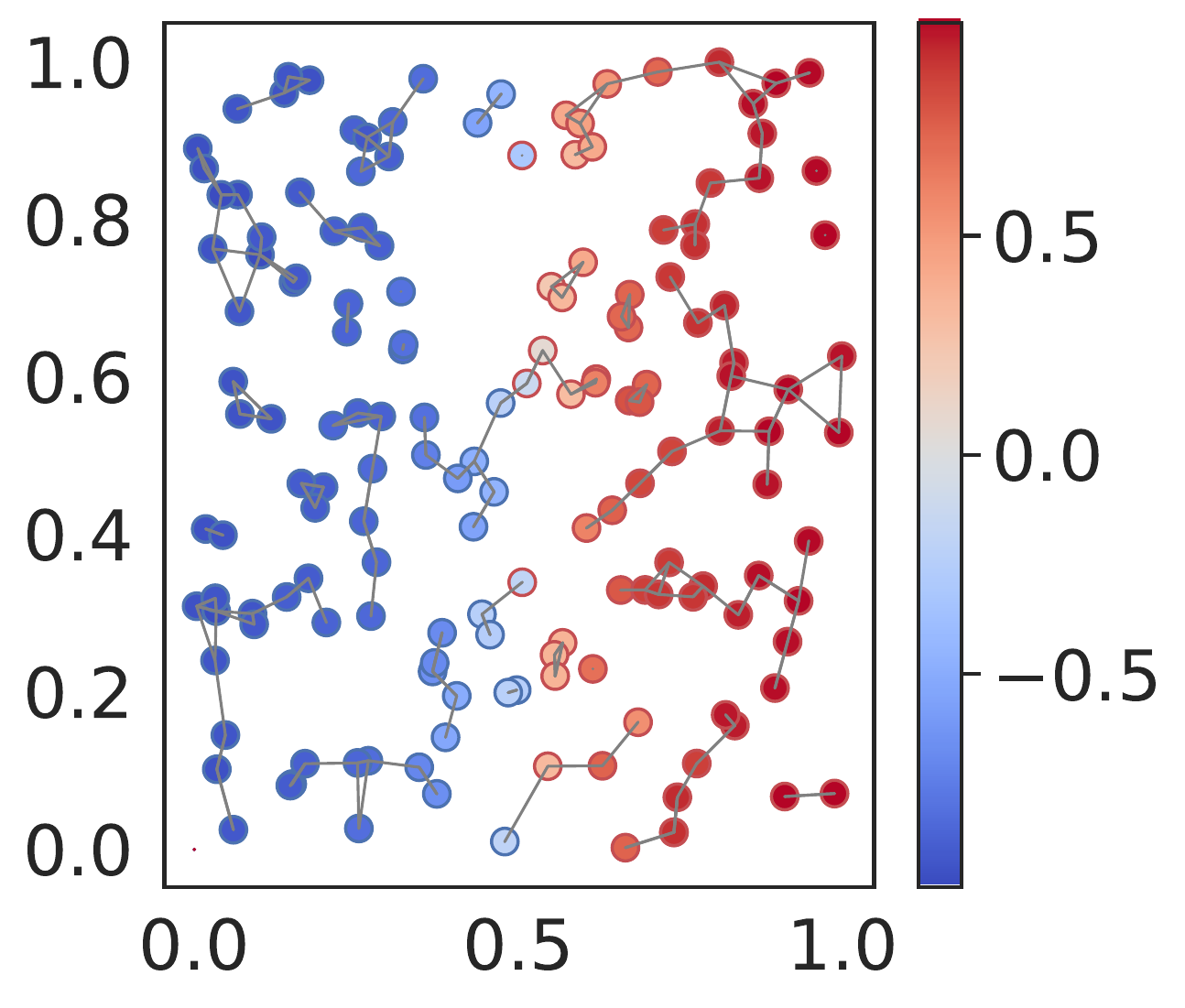}	
			\label{fig:2e}
		\end{minipage}
		}
		\vspace{-10pt}
	\caption{Linearly separable dataset. The accuracy of models (a) to (d) is 0.97, 0.51, 0.96, and 0.98 respectively. The face color of each node shows its prediction score and the border color shows the ground-truth label. Grey lines connect the nearest neighbours.
	Since the perturbation is large ($\rho=2.5$), AWP fails. The proposed weighting and truncation mitigate the vanishing-gradient issue for the same $\rho$.
	}
	\label{fig:2}
	\vspace{-15pt}
\end{figure*}

Recall that the AWP objective is the unweighted combination of the regular loss function $L({\bm{\theta}})$ and the sharpness term $\max_{{\bm{\delta}\leq\rho}} [L({\bm{\theta}}+{\bm{\delta}})-L({\bm{\theta}})]$ (Sec. \ref{sec:relatedw}). The weight perturbation in this term can lead to vanishing gradients as we discussed in Sec. \ref{sec:vanishing_gradient}. Therefore, another way to deal with this issue is to assign a smaller weight $\lambda$ to the sharpness term in the AWP objective. The weighted combination is $[\lambda\max_{{\bm{\delta}\leq\rho}} [L({\bm{\theta}}+{\bm{\delta}})-L({\bm{\theta}})] + L({\bm{\theta}})] = [\lambda\max_{{\bm{\delta}\leq\rho}} L({\bm{\theta}}+{\bm{\delta}}) + (1-\lambda)L({\bm{\theta}})]$.
\begin{definition}(Weighted AWP) Given a weight $\lambda\in[0,1]$ the Weighted AWP objective is
\begin{equation}
\label{eq:weighted_awp}
\min_{{\bm{\theta}}}[\lambda L_\textup{train}({\bm{\theta}}+\hat{\bm{\delta}}^*({\bm{\theta}});\bm{A},\bm{X})+(1-\lambda)L_\textup{train}({\bm{\theta}};\bm{A},\bm{X})]
    \vspace{-9pt}
\end{equation}
\end{definition}

We compare these two improvements with AWP and natural training on a linearly separable dataset using the same setup as in Sec. \ref{sec:vanishing_gradient}. 
%
Figure \ref{fig:2} illustrates the trained models with $\rho = 2.5$. In  Figure \ref{fig:2b} we can see that the model with AWP objective suffers from vanishing gradients and it fails to learn anything useful.
The models with Truncated AWP\footnote{
    In Figure \ref{fig:2c} we
    perturb only the first-layer, i.e. $\bm\theta^\textup{(awp)}=\bm{W}_1$ (first layer weights) and $\bm\theta^\textup{(normal)}=\bm{W}_2$ (last layer weights).
    Perturbing only the second layer instead performs similarly.
    } 
    (Figure \ref{fig:2c}) and Weighted AWP (Figure \ref{fig:2d}) mitigate this issue, which is also evident in their learning curves (Figure \ref{fig:awpknnloss}), and have relatively good performance (96\% and 98\% accuracy respectively). %
Compared to the vanilla model (Figure \ref{fig:2a}), they both have a significantly smoother decision boundary.

To tackle the vanishing-gradient issue better, we combine Truncated AWP and Weighted AWP, into a Weighted Truncated Adversarial Weight Perturbation (WT-AWP). The details of WT-AWP are shown in Algorithm \ref{alg:2} (description in Sec. \ref{sec:despalg}). WT-AWP has two important parameters $\lambda$ and $\rho$. We study how they influence the model performance in Sec. \ref{sec:ablation}.

\begin{algorithm}[t]
\caption{WT-AWP: Weighted Truncated Adversarial \\ Weight Perturbation}
\label{alg:2}
\textbf{Input}: Graph $G=(\bm{A},\bm{X})$; model parameters ${\bm{\theta}} = [{\bm{\theta}}^\text{(awp)}; {\bm{\theta}}^\text{(normal)}]$ with and without AWP; number of epochs $T$; loss function $L_\text{train}$; perturbation strength $\rho$, AWP weight $\lambda$; learning rate $\alpha$.
\begin{algorithmic}[1] 
\STATE Initialize weight ${\bm{\theta}}_0$;.
\FOR{$t\in$ 1:T}
\STATE Compute the loss for training nodes: $L_\text{train}({\bm{\theta}}_{t-1};\bm{A},\bm{X})$\\
\STATE	Compute the approximating weight perturbation for ${\bm{\theta}}_{t-1}^\textup{(awp)}$: $\hat{\bm{\delta}}^*({\bm{\theta}}_{t-1}^\text{(awp)})$ via Eq. \ref{eq:est}\\
\STATE	Compute the approximating gradient for ${\bm{\theta}}$:\\ 
	$\begin{aligned}
	\bm{g} = &\lambda\nabla_{\bm{\theta}} L_\text{train}({\bm{\theta}};\bm{A},\bm{X})|_{{\bm{\theta}}_{t-1}+[\hat{{\bm{\delta}}}^*({\bm{\theta}}_{t-1}^\textup{(awp)}),0]} \\ &+(1-\lambda)\nabla_{\bm{\theta}} L_\text{train}({\bm{\theta}};\bm{A},\bm{X})|_{{\bm{\theta}}_{t-1}}
	\end{aligned}
	$\\
\STATE	Update the weight via ${\bm{\theta}}_t = {\bm{\theta}}_{t-1}-\alpha \bm{g}$ 
\ENDFOR
\STATE \textbf{return} ${\bm{\theta}}_T$
\end{algorithmic}
\end{algorithm}



\section{Experimental Results}\label{sec:experiment}
\textbf{Setup.} 
We conduct comprehensive experiments to show the effect of WT-AWP  on the natural and robustness performance of different GNNs for both node classification and graph classification tasks.
We utilize the open-source libraries \emph{Pytorch-Geometric}
~\cite{Fey/Lenssen/2019} 
and \emph{Deep-Robust}
~\cite{li2020deeprobust} 
to evaluate clean and robust node classification performance respectively. To achieve fair comparison we keep the same training settings for all models. We report the mean and standard deviation over 20 different train/val/test splits and 10 random weight initializations. See Sec. \ref{sec:training setup} for further details and hyperparameters.

\noindent\textbf{Datasets.} We use three benchmark datasets, including two citation networks, Cora and Citeseer \cite{sen2008collective}, and one blog dataset Polblogs \cite{adamic2005political}. We treat all graphs as undirected and only select the largest connected component (more details and statistics in Sec. \ref{sec:data_and_setup}).

\begin{table}[]
\caption{Clean accuracy comparison. We report the average and the standard deviation across 200 experiments per model (20 random splits $\times$ 10 random initializations). WT-AWP consistently outperform the standard models on all benchmarks. The improvements are statistically significant according to a two-sided t-test at a significance level of  $p<0.001$.}
\vspace{-10pt}
\centering
\scalebox{0.87}{
\begin{tabular}{cccc}
\hline
\hline
Approachs    & Cora                  & Citeseer              & Polblogs              \\ \hline
GCN          & 84.14 ± 0.61          & 73.44 ± 1.35          & 95.04 ± 0.66          \\
GCN+WT-AWP   & 85.16 ± 0.44          & 74.48 ± 1.04          & 95.26 ± 0.51          \\ \hline
GAT          & 84.13 ± 0.79          & 73.71 ± 1.23          & 94.93 ± 0.51          \\
GAT+WT-AWP   & 85.13 ± 0.51          & 74.73 ± 1.07          & 95.12 ± 0.48          \\ \hline
PPNP        & 85.56 ± 0.46          & 74.50 ± 1.06          & 95.18 ± 0.42          \\
PPNP+WT-AWP & \textbf{86.13 ± 0.43} & \textbf{75.64 ± 0.95} & \textbf{95.36 ± 0.37} \\ \hline
\end{tabular}
}
\vspace{-15pt}
\label{tab:natacc}
\end{table}

\begin{figure}[t]
	\subfigure[Cora adj. matrix]{
		\begin{minipage}[t]{.45\linewidth}
			\centering
			\includegraphics[height=3cm]{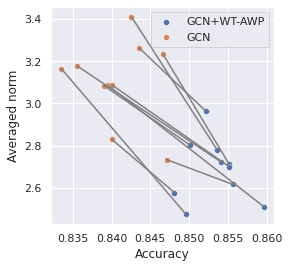}	
			\label{fig:lipadjcora}
		\end{minipage}
	}
	\subfigure[Cora node feat.]{
		\begin{minipage}[t]{.45\linewidth}
			\centering
			\includegraphics[height=3cm]{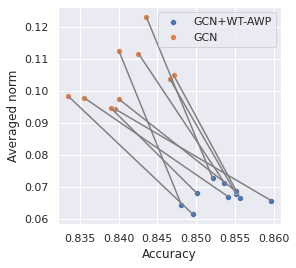}\vspace{-1em}	
			\label{fig:lipfeatcora}
		\end{minipage}
	}\\
	\subfigure[Citeseer adj. matrix]{
		\begin{minipage}[t]{.45\linewidth}
			\centering
			\includegraphics[height=3cm]{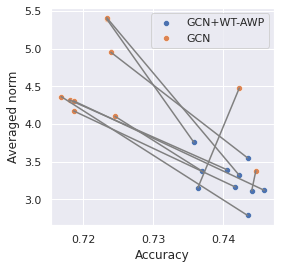}	
			\label{fig:lipciteadj}
		\end{minipage}
	}
	\subfigure[Citeseer node feat.]{
		\begin{minipage}[t]{.45\linewidth}
			\centering
			\includegraphics[height=3cm]{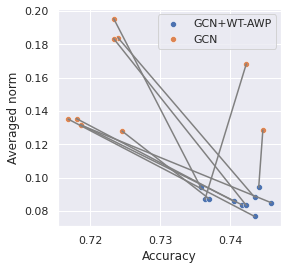}	
			\label{fig:lipfeatcite}
		\end{minipage}
	}
	\vspace{-5pt}
	\caption{Comparison of the averaged gradient norm w.r.t. the adjacency matrix and the node features for GCN models with and without WT-AWP on Cora and Citeseer. Each connected pair of points refers to a GCN and a GCN+WT-AWP model trained with the same data split and initialization.}
	\label{fig:lipcora}
	\vspace{-15pt}
\end{figure}

\begin{figure}[]
		\centering
		\includegraphics[height=3.5cm]{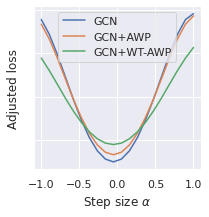}	
		\vspace{-0.5cm}
		\caption{Loss landscape for different models.}
		\label{fig:lossland}
		\vspace{-25pt}
\end{figure}

\noindent\textbf{Baseline models and attacks.} 
We aim to evaluate the impact of our WT-AWP on natural and robust node classification tasks.
We train three vanilla GNNs: GCN \cite{kipf2016semi}, GAT \cite{velivckovic2017graph}, and PPNP \cite{klicpera2018predict}, and four graph defense methods: 
RGCN \cite{zhu2019robust}\footnote{Note, we cannot apply WT-AWP to RGCN as the weights are modeled as (Gaussian) distributions.}, 
GCNJaccard \cite{wu2019adversarial}, GCNSVD \cite{entezari2020all}, and SimpleGCN \cite{jin2021node}. For detailed baseline descriptions see Sec. \ref{sec:baselinemodel}.

To generate the adversarial perturbations, we apply three methods including: DICE \cite{waniek2018hiding}, PGD \cite{xu2019topology}, and Metattack \cite{zugner2019adversarial}. For a discussion of the attacks see Sec. \ref{sec:graphattacks}.

\noindent\textbf{Certified robustness.}
We obtain provable guarantees for our models using a black-box (sparse) randomized smoothing certificate \cite{bojchevski2020efficient}.
We report the the certified accuracy, i.e. the percentage of nodes guaranteed to be correctly classified, given an adversary that can delete up to $r_d$ edges or add up to $r_a$ edges to graph (similarly for the node features). See Sec. \ref{sec:rand_smooth_details} for details.

\noindent\textbf{Settings for WT-AWP.} All baseline models have a 2-layer structure. When applying the WT-AWP objective, we only perform weight perturbation on the \emph{first} layer i.e. we assign $\bm\theta^\text{(awp)}=\bm{W}_1$ (the first layer) and $\bm\theta^\text{(normal)}=\bm{W}_2$ (the last layer). For generating the weight perturbation we use a 1-step PGD as discussed in Sec. \ref{sec:gnnawp}.
In the ablation study Sec. \ref{sec:ablation} we also apply 5-step PGD to generate weight perturbation, in which we utilize an SGD optimizer with learning rate 0.2 and update the perturbation for 5 steps. In the end we project the perturbation on the $l_2$ ball $B(\rho({\bm{\theta}}))$.\looseness=-1


\begin{table*}[t]
\centering

\caption{Robust accuracy under PGD and Metattack poisoning attacks, with a 5\% adversarial budget. We report the average and the standard deviation across 200 experiments per model (20 random splits $\times$ 10 random initializations). Our WT-AWP loss improves over all (vanilla and robust) baselines. All results expect the one marked with * are statistically significant at $p<0.05$ according to a t-test.
}
\vspace{-10pt}
\scalebox{0.7}{
\renewcommand{\arraystretch}{1.2}
\begin{tabular}{cccccccccc}
\hline\hline
                       & \multicolumn{3}{c}{Natural Acc}                                       & \multicolumn{3}{c}{Acc with 5\% PGDattack}                            & \multicolumn{3}{c}{Acc with 5\% Metattack}                            \\ \hline
Models & Cora                  & Citeseer              & Polblogs              & Cora                  & Citeseer              & Polblogs*              & Cora                  & Citeseer              & Polblogs              \\ \hline 
GCN                    & 83.73 ± 0.71          & 73.03 ± 1.19          & 95.06 ± 0.68          & 81.26 ± 1.27          & 72.04 ± 1.60          & 85.18 ± 2.63          & 78.61 ± 1.66          & 69.20 ± 1.93          & 79.74 ± 1.05          \\ 
+WT-AWP             & \textbf{84.66 ± 0.53} & 74.01 ± 1.11          & \textbf{95.20 ± 0.61} & 82.66 ± 1.07          & 73.73 ± 1.23          & \textbf{85.73 ± 4.17} & 79.05 ± 1.73          & 70.50 ± 1.65          & 80.72 ± 1.25          \\ \hline 
GCNJaccard             & 82.42 ± 0.73          & 73.09 ± 1.20          & N/A                   & 80.65 ± 1.14          & 72.05 ± 1.76          & N/A                   & 78.96 ± 1.54          & 69.62 ± 1.87          & N/A                   \\ 
+WT-AWP      & 83.55 ± 0.60          & 74.10 ± 1.04          & N/A                   & 82.12 ± 0.91          & 73.85 ± 1.38          & N/A                   & \textbf{80.23 ± 1.38} & 71.22 ± 1.44          & N/A                   \\ \hline
SimPGCN                & 82.99 ± 0.68          & 74.05 ± 1.28          & 94.67 ± 0.95          & 80.71 ± 1.33          & 73.61 ± 1.39          & 82.42 ± 3.14          & 78.60 ± 1.81          & 72.52 ± 1.72          & 76.66 ± 1.80          \\ 
+WT-AWP         & 83.37 ± 0.74          & \textbf{74.26 ± 1.09} & 94.85 ± 0.91          & \textbf{83.49 ± 0.78} & \textbf{74.43 ± 1.14} & 82.68 ± 4.82          & 79.76 ± 1.76          & \textbf{72.95 ± 1.43} & 77.68 ± 2.41          \\ \hline
GCNSVD                 & 77.63 ± 0.63          & 68.57 ± 1.54          & 94.08 ± 0.59          & 76.83 ± 1.42          & 68.08 ± 1.98          & 82.84 ± 3.05          & 76.28 ± 1.15          & 67.34 ± 1.93          & 91.76 ± 1.19          \\ 
+WT-AWP          & 79.05 ± 0.58          & 71.12 ± 1.42          & 94.13 ± 0.59          & 78.50 ± 0.89          & 71.43 ± 1.46          & 82.97 ± 3.57          & 77.61 ± 1.08          & 70.65 ± 1.28          & \textbf{92.28 ± 0.98} \\ \hline 
RGCN                   & 83.29 ± 0.63          & 71.69 ± 1.35          & 95.15 ± 0.46          & 78.47 ± 1.10          & 68.81 ± 2.32          & 85.62 ± 1.51          & 77.70 ± 1.69          & 69.05 ± 1.90          & 79.48 ± 1.16          \\ \hline
\end{tabular}
}\label{tab:poison}
\vspace{-8pt}
\end{table*}
\begin{table*}[t]
\centering
\caption{Robust accuracy under evasion attacks of different strength. We report the average and the standard deviation across 200 experiments per model (20 random splits $\times$ 10 random initializations). Our WT-AWP loss always improves the robustness of the baselines. All results expect the one marked with * are statistically significant at $p<0.05$ according to a two-sided t-test.}
\vspace{-10pt}
\scalebox{0.7}{
\renewcommand{\arraystretch}{1.2}
\begin{tabular}{cccccccc}
\hline\hline
\multicolumn{1}{l}{}        & \multicolumn{1}{l}{Perturbation strength} & \multicolumn{3}{c}{5\%}                                               & \multicolumn{3}{c}{10\%}                                              \\ \hline
\multicolumn{1}{l}{Attacks} & Models                                 & Cora                  & Citeseer              & Polblogs              & Cora                  & Citeseer              & Polblogs              \\ \hline
\multirow{2}{*}{DICE}       & GCN                                       & 82.83 ± 0.87          & 71.85 ± 1.31          & 91.27 ± 0.98          & 81.87 ± 0.94          & 71.17 ± 1.50          & 87.47 ± 1.17          \\
                            & +WT-AWP                                    & \textbf{84.01 ± 0.59} & \textbf{73.84 ± 1.10} & \textbf{91.45 ± 0.86}* & \textbf{82.93 ± 0.64} & \textbf{73.14 ± 1.25} & \textbf{87.70 ± 0.97} \\ \hline
\multirow{2}{*}{PGD}        & GCN                                       & 79.92 ± 0.62          & 70.50 ± 1.35          & 79.41 ± 0.76          & 77.17 ± 0.74          & 68.49 ± 1.39          & 72.90 ± 0.73          \\
                            & +WT-AWP                                    & \textbf{81.00 ± 0.56} & \textbf{70.69 ± 1.45}* & \textbf{80.70 ± 0.90} & \textbf{77.87 ± 0.64} & \textbf{68.96 ± 1.30} & \textbf{75.11 ± 1.03} \\ \hline
\end{tabular}
}\label{tab:evasion}
\vspace{-10pt}
\end{table*}
\subsection{Clean Accuracy}\label{sec:cleanaccuracy}
We evaluate the clean accuracy of node classification tasks for different GNNs and benchmarks. The baselines include GCN, GAT, and PPNP . We use a 2-layer structure (input-hidden-output) for these three models. For GCN and PPNP, the hidden dimensionality is 64; for GAT, we use 8 heads with size 8. We choose $K=10,\alpha=0.1$ in PPNP. We also find that the hyperparameters $(\lambda,\rho)$ of WT-AWP are more related to the dataset than the backbone models. We use $(\lambda=0.7,\rho=1)$ for all three baseline models on Cora, $(\lambda=0.7,\rho=2.5)$ on Citeseer, and $(\lambda=0.3,\rho=1)$ for GCN,  $(\lambda=0.3,\rho=2)$ for GAT and PPNP on Polblogs.
Table \ref{tab:natacc} shows our results, WT-AWP clearly improves the accuracy of all baseline models, while having smaller standard deviations. Note, we do not claim that these models are state of the art, but rather that WT-AWP provides consistent and statistically significant (two-sided t-test, $p<0.001$) improvements over the baseline models. These results support our claim that WT-AWP finds local minima with better generalization properties.

\subsection{Models Trained with WT-AWP are Smoother}
\label{sec:average_gradient}
To estimate the smoothness of the loss landscape around the adjacency matrix $\bm{A}$ and the node attributes $\bm{X}$, we compute the average norm of the gradient of $L_\text{train}({\bm{\theta}};\bm{A},\bm{X})$ \emph{w.r.t.} $\bm{A}$ and $\bm{X}$. We compare a vanilla GCN model with GCN+WT-AWP ($\lambda = 0.5, \rho = 1$) model on Cora and Citeseer. We train 10 models with different random initializations. For each model we randomly sample 100 noisy inputs around $\bm{A}$ and $\bm{X}$, and we average the gradient norm for these noisy inputs. 
When comparing models trained with and without WT-AWP, we keep everything else fixed, including the random initialization, to isolate the effect of WT-AWP.
In Figure \ref{fig:lipcora}, we can observe that in most cases  (37 out of 40) the models trained with WT-AWP have both better accuracy and smaller average gradient norm, \emph{i.e.} are smoother. As we show in Sec. \ref{sec:poisoning} and Sec. \ref{sec:evasion} this consequently improves their robustness to adversarial input perturbations.


\subsection{Visualization of Loss Landscape}
\label{sec:loss_landspace}

We train GCN, GCN+AWP ($\rho=0.1$) and GCN+WT-AWP ($\lambda = 0.5, \rho = 0.5$) models with the same initialization, and we compare their loss landscapes. The accuracy is 83.55\% for GCN, 84.21\% for GCN+AWP, and 85.51\% for GCN+WT-AWP. 
Similar to \citet{stutz2021relating}, Figure \ref{fig:lossland} shows the loss landscape in a randomly chosen direction $\bm{u}$ in weight space, \emph{i.e.} we plot $L_\text{train}(\bm{\theta}+\alpha \cdot \bm{u}; \bm{A}, \bm{X})$ for different steps $\alpha$.
We generate 10 random directions $\bm{u}$ and show the average loss.
The loss landscape of GCN+AWP is slightly flatter than the vanilla GCN, because of the small perturbation strength $\rho=0.1$.  GCN+WT-AWP is flatter than both of them (and more accurate) due to the larger perturbation strength $\rho = 0.5$. This experiment provides further evidence for the effectiveness of WT-AWP.

\begin{table*}[t]
\centering
\vspace{-6pt}
\caption{Hyperparameter sensitivity study for $\lambda$ and $\rho$ on the Cora dataset for a GCN base model.}
\vspace{-10pt}
\label{tab:ablation}
\scalebox{0.8}{
\begin{tabular}{c|cccccc}
\hline
\hline
WT-AWP& $\rho = 0.05$ & $\rho = 0.1$ & $\rho = 0.5$ & $\rho = 1$   & $\rho = 2.5$ & $\rho = 5$   \\ \hline
$\lambda = 0.1$                                                                                                                     & 84.15 ± 0.60  & 84.15 ± 0.61 & 84.51 ± 0.48 & 84.58 ± 0.52 & 84.50 ± 0.51 & 84.54 ± 0.49 \\ 
$\lambda = 0.3$                                                                                                                     & 84.10 ± 0.62  & 84.13 ± 0.58 & 84.76 ± 0.51 & 84.91 ± 0.46 & 84.77 ± 0.46 & 84.64 ± 0.47 \\ 
$\lambda = 0.5$                                                                                                                     & 84.11 ± 0.64  & 84.09 ± 0.61 & 84.93 ± 0.49 & 85.06 ± 0.49 & 84.94 ± 0.45 & 84.67 ± 0.49 \\ 
$\lambda= 0.7$                                                                                                                      & 84.13 ± 0.59  & 84.15 ± 0.64 & 85.00 ± 0.46 & \textbf{85.16 ± 0.44} & 84.99 ± 0.49 & 84.66 ± 0.49 \\ 
$\lambda = 1.0$                                                                                                                       & 84.12 ± 0.69  & 84.23 ± 0.64 & 82.45 ± 1.98 & 60.29 ± 1.94 & 29.51 ± 0.91 & 29.19 ± 0.13 \\ \hline 
 AWP                                                                                                                        & 84.16 ± 0.68  & 84.23 ± 0.68 & 41.19 ± 1.23 & 29.18 ± 0.07 & 29.18 ± 0.02 & 29.18 ± 0.02\\
 W-AWP                                                                                                                       & 84.12 ± 0.66  & 84.20 ± 0.66 & 84.63 ± 0.51 & 84.32 ± 0.65 & 83.98 ± 0.93 & 83.62 ± 1.27\\
\hline
\end{tabular}

}
\end{table*}

\begin{table*}[t]
\centering
\vspace{-5pt}
\caption{Ablation study with $\lambda$ and $\rho$ on WT-AWP, where the weight perturbation is calculated with 5-step PGD. The backbone model is GCN and the benchmark is Cora. We observe no significant improvement compared to the computationally less expensive 1-step PGD.}
\vspace{-10pt}
\label{tab:ablation2}
\scalebox{0.8}{
\begin{tabular}{c|cccccc}
\hline
\hline
WT-AWP (5 step) & $\rho = 0.05$ & $\rho = 0.1$ & $\rho = 0.5$ & $\rho = 1$            & $\rho = 2.5$ & $\rho = 5$   \\ \hline
$\lambda = 0.1$ & 84.19 ± 0.60  & 84.17 ± 0.59 & 84.45 ± 0.51 & 84.50 ± 0.50          & 84.39 ± 0.52 & 84.41 ± 0.54 \\
$\lambda = 0.3$ & 84.12 ± 0.58  & 84.15 ± 0.63 & 84.65 ± 0.54 & 84.81 ± 0.47          & 84.70 ± 0.50 & 84.55 ± 0.55 \\
$\lambda = 0.5$ & 84.10 ± 0.59  & 84.11 ± 0.62 & 84.77 ± 0.53 & 84.90 ± 0.50          & 84.82 ± 0.47 & 84.64 ± 0.52 \\
$\lambda= 0.7$  & 84.12 ± 0.61  & 84.11 ± 0.63 & 84.86 ± 0.49 & \textbf{84.99 ± 0.48} & 84.89 ± 0.51 & 84.64 ± 0.52 \\
$\lambda = 1.0$   & 84.11 ± 0.62  & 84.18 ± 0.63 & 72.18 ± 1.48 & 32.55 ± 6.80          & 29.18 ± 0.03 & 29.18 ± 0.00 \\ \hline
\end{tabular}
}
\vspace{-14pt}
\end{table*}
\subsection{Robust Accuracy with Poisoning Attacks}\label{sec:poisoning}
Next we show that our WT-AWP can improve existing defense methods against graph poisoning attacks. We select two poisoning attacks: PGD and Metattack \cite{zugner2019adversarial}, with a 5\% adversarial budget. The baseline models are vanilla GCN, and three GCN-based graph-defense models: GCNJaccard, GCNSVD, and SimpleGCN. 
For all attack and defense methods, we apply the default hyperparameter settings in \cite{li2020deeprobust}, which re-implements the corresponding models with the same hyperparameters as the original works. We use Cora, Citeseer, and Polblogs as the benchmark datasets. Note that GCNJaccard does not work on Polblogs as it requires node features. Table \ref{tab:papoi} in the appendix shows the hyperparameters $(\lambda,\rho)$ we select for all WT-AWP models.

As we can see in Table \ref{tab:poison}, none of the defense methods have dominant performance across benchmarks. More importantly, our WT-AWP consistently improves the robust accuracy for both vanilla and robust models. 
We also evaluate the models against the DICE poisoning attack in Sec. \ref{sec:poisoningdice}, and again the results demonstrate that WT-AWP adds meaningful improvement over the baselines.

\subsection{Robust Accuracy with Evasion Attacks}
\label{sec:evasion}
Next we show that WT-AWP also improves existing defense methods against evasion attacks. We select two evasion attacks, DICE and PGD, with perturbation strengths of 5\% and 10\%.
The baseline model is GCN and we perform experiments on three benchmarks: Cora, Citeseer, and Polblogs.
For the PGD attack the hyperparameters $(\lambda,\rho)$ are (0.5, 0.5) for all datasets. For the DICE attack we use (0.5, 0.5) for Cora, (0.7, 2) for Citeseer, and (0.3, 1) for Polblogs.
Table \ref{tab:evasion} shows the experimental results. WT-AWP again meaningfully improves the robustness of GCN under both PGD and DICE evasion attacks for all perturbation strengths.

\begin{figure}[!b]
\centering
	\subfigure[Node feature]{
		\begin{minipage}[t]{0.45\linewidth}
			\centering
			\includegraphics[height=2.5cm]{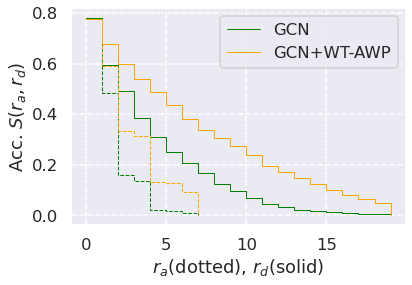}	
		\end{minipage}
	}
	\subfigure[Graph structure]{
		\begin{minipage}[t]{0.45\linewidth}
			\centering
			\includegraphics[height=2.5cm]{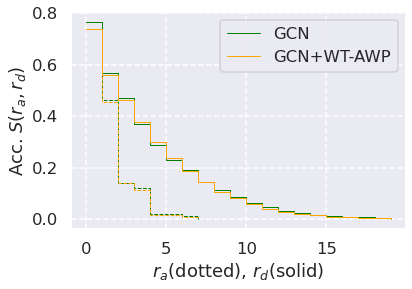}	
		\end{minipage}
	}
	\vspace{-10pt}
	\caption{Robustness guarantees on Cora, where $r_a$ is the certified radius -- maximum number of adversarial additions (and $r_d$ for deletions). For perturbations to the node features WT-AWP significantly improves the certified accuracy, i.e. the number of nodes guaranteed not to change their prediction, for all certified radii.}
	\label{fig:certcora}
	\vspace{-15pt}
\end{figure}

\subsection{Ablation and Hyperparameter Sensitivity Study}\label{sec:ablation}
We compare the performance of GCN+WT-AWP on the Cora dataset for different $\lambda$ and $\rho$ values. We also compare WT-AWP with AWP under different perturbation strengths $\rho$. Table \ref{tab:ablation} lists the results. The accuracy of GCN+WT-AWP first increases with $\lambda$ and $\rho$ and then slightly decreases. 
Truncated AWP is a special case for $\lambda=1$ (since the $(1-\lambda)$ term disappears in Eq. \ref{eq:weighted_awp}) and it does not perform well, especially for larger $\rho$.
Similarly, WT-AWP outperforms the vanilla AWP that suffers from the vanishing-gradient issue.
Weighted but not truncated AWP with $\lambda = 0.5$ (last row) is also worse than WT-AWP, although in general weighting seems to be more important than truncation.
These results justify the decision to combine weighting and truncation.

We also generate perturbations as in Eq. \ref{eq:est} but with multi-step PGD. We repeat the above experiment with applying 5-step PGD for WT-AWP. As shown in Table \ref{tab:ablation2}, the performance of 5-step WT-AWP is similar to the 1-step WT-AWP, the accuracy of  both models first increases with $\lambda$ and $\rho$, and then decreases. The optimal hyperparameters $(\lambda,\rho)$ are $\rho=1,\lambda=0.7$. 
Since 5-step PGD offers no benefits and 1-step PGD is computationally less expensive, we suggest this as the default setting when applying WT-AWP.


\vspace{-3pt}
\subsection{Certified Robustness}
In this subsection, we measure the certified robustness of GCN and GCN+WT-AWP on the Cora dataset with sparse randomized smoothing \cite{bojchevski2020efficient}.  We use $\lambda = 0.5, \rho=1$ as the hyperparameters for the WT-AWP models.
We plot the certified accuracy $S(r_a,r_d)$ for different addition $r_a$ and deletion $r_d$ radii. In Figure \ref{fig:certcora}, we see that compared to vanilla GCN training, our WT-AWP loss increases the certified accuracy 
\emph{w.r.t.} feature perturbations for all radii, while maintaining comparable performance when certifying perturbations of the graph structure.
For additional results see Sec. \ref{sec:certcite}.

\vspace{-5pt}
\section{Conclusion}
We proposed a new adversarial weight perturbation method, WT-AWP, and we evaluated it on graph neural networks. We showed that our WT-AWP can improve the regularization of GNNs by finding flat local minima. We conducted extensive experiments to validate our method. In all empirical results, WT-AWP consistently improves the performance of GNNs on a wide range of graph learning tasks including node classification, graph defense, and graph classification. Further exploring the connections between flat minima and generalization in GNNs is a promising research direction.

\section*{Acknowledgements}
This work was partially supported by NSF IIS 1838627, 1837956, 1956002, 2211492, CNS 2213701, CCF 2217003, DBI 2225775.
\bibliography{aaai23.bib}

\newpage
\appendix
\onecolumn
\begin{center}
{\Large Supplementary Material}
\end{center}
\section{Proof of Theorem 1}\label{sec:genebound}

\textbf{Theorem 1.} \textit{(generalization bound) Assume there exists a $\bm{\Sigma}$ such that $L_{\textup{all}}(\bm{\theta};\bm{A},\bm{X})\leq\mathbb{E}_{\bm{z}\sim\mathcal{N}(\bm{0},\bm{\Sigma})}[L_{\textup{all}}(\bm{\theta}+\bm{z};\bm{A},\bm{X})]$. For any set of training nodes $\mathcal{V}_\textup{train}$ from $\mathcal{V}_\textup{all}$, $\forall m\geq\sqrt{d}$, with probability at least $1-\delta$, we have}
\begin{equation}
\begin{aligned}
    L_{\textup{all}}(\bm{\theta};\bm{A},\bm{X})\leq\max_{||\bm{\delta}||_2\leq \rho}[L_{\textup{train}}(\bm{\theta}+\bm{\delta};\bm{A},\bm{X})]+(\frac{m^2}{d}e^{1-\frac{m^2}{d}})^{d/2}\\
+\frac{1}{\sqrt{N_0}}\left(\frac{1}{2}\left[1+d\log(1+\frac{m^2||\bm{\theta}||_2^2}{d\rho^2})\right]+\ln\frac{3}{\delta}+\frac{1}{4}+\Theta(K \cdot \epsilon_\textup{all})\right).
\end{aligned}
\end{equation}
\textit{where $d$ is the number of parameters in the GNN, $K$ is the number of groundtruth labels, $\epsilon_\textup{all}$ is a fixed constant w.r.t. $\mathcal{V}_\textup{all}$, $N_0$ is the volume of $\mathcal{V}_\textup{train}$.
}

We have the same assumption as in \citet{foret2020sharpness}, $L_{\text{all}}(\bm{\theta};\bm{A},\bm{X})\leq\mathbb{E}_{\bm{z}\sim\mathcal{N}(\bm{0},\bm{\Sigma})}[L_{\text{all}}(\bm{\theta}+\bm{z};\bm{A},\bm{X})]$, which means that adding Gaussian perturbation  should not decrease the test error.
\begin{proof}
Our proof is motivated by the subgroup generation bound on node classification tasks \cite{ma2021subgroup} and the intuition of Theorem 1 in \cite{foret2020sharpness}.

\begin{lemma}\label{lm:1}(PAC-Bayes bound for node classification tasks \cite{ma2021subgroup}) For any set of training nodes $\mathcal{V}_\textup{train}$ from $\mathcal{V}_\textup{all}$, for any subgroup of nodes $\mathcal{V}_{m}\subset\mathcal{V}_\textup{all}$, 
for any prior distribution $\mathcal{P}$, with probability at least $1-\delta$, for any distribution $\mathcal{Q}$ we have
\begin{equation}
    \mathbb{E}_{\bm{\theta}\sim\mathcal{Q}}[L_{\text{m}}(\bm{\theta};\bm{A},\bm{X})]\leq  \mathbb{E}_{\bm{\theta}\sim\mathcal{Q}}[L_{\textup{train}}(\bm{\theta};\bm{A},\bm{X})]+\frac{1}{\sqrt{N_0}}(\mathbb{D}_{KL}(\mathcal{Q}||\mathcal{P})+\ln\frac{3}{\delta}+\frac{1}{4}+\Theta(K\epsilon_m))
\end{equation}
where $N_0$ is the volume of the training set $\mathcal{V}_\textup{train}$, $K$ is the total number of classes, $\epsilon_m$ is a constant depend on the subgroup $\mathcal{V}_{m}$.
\end{lemma}
If we take $\mathcal{V}_{m}=\mathcal{V}_\textup{all}$ in Lemma~\ref{lm:1}, we have
\begin{equation}
    \mathbb{E}_{\bm{\theta}\sim\mathcal{Q}}[L_{\textup{all}}(\bm{\theta};\bm{A},\bm{X})]\leq  
    \mathbb{E}_{\bm{\theta}\sim\mathcal{Q}}[L_{\textup{train}}(\bm{\theta};\bm{A},\bm{X})]
    +\frac{1}{\sqrt{N_0}}(\mathbb{D}_{KL}(\mathcal{Q}||\mathcal{P})+\ln\frac{3}{\delta}+\frac{1}{4}+\Theta(K\epsilon_\textup{all}))
    \label{eq:pacgnn}
\end{equation}

Assume both $\mathcal{P}$ and $\mathcal{Q}$ are Gaussian distributions with isotropic covariance matrices, i.e. $\mathcal{P}\sim\mathcal{N}(\bm{\mu}_{p},{\sigma}_{p}^2\bm{I}_d\})$, $\mathcal{Q}\sim\mathcal{N}(\bm{\mu}_{q},{\sigma}_{q}^2\bm{I}_d\})$, where $d$ is the dimension of $\bm{\theta}$, we have

\begin{equation}
\mathbb{D}_{KL}(\mathcal{Q}||\mathcal{P}) = \frac{1}{2}\left[d\log\frac{\sigma_{p}^2}{\sigma_{q}^2}-d+d\frac{\sigma_{q}^2}{\sigma_{p}^2}+||\frac{{\bm{\mu}}_{p}-{\bm{\mu}}_{q}}{\sigma_p}||_2^2 \right],
\end{equation}
Take $\bm{\mu}_{q} = \bm{\theta}, \bm{\mu}_{p} = \bm{0}$, we expect the KL-divergence $\mathbb{D}_{KL}(\mathcal{Q}||\mathcal{P})$ to be as small as possible w.r.t. $\sigma_{p}$. 
\begin{lemma}\label{lm:2}\cite{foret2020sharpness}
Take ${\mu}_{q} = {\theta}, {\mu}_{p} = {0}$. There exist pre-defined ${\sigma}_{p}$ such that 
\begin{equation}
\mathbb{D}_{KL}(\mathcal{Q}||\mathcal{P})\leq \frac{1}{2}\left[1+d\log(1+\frac{||\bm{\theta}||_2^2}{d\sigma_q^2})\right]
\end{equation}
\end{lemma}
 
 Thus we have the generalization bound 
 \begin{equation}
 \begin{aligned}
    &\mathbb{E}_{\theta\sim\mathcal{Q}}[L_{\text{all}}(\bm{\theta};\bm{A},\bm{X})]\leq\mathbb{E}_{\bm{\theta}\sim\mathcal{Q}}[L_{\text{train}}(\bm{\theta};\bm{A},\bm{X})]\\
    &+\frac{1}{\sqrt{N_0}}\left(\frac{1}{2}\left[1+d\log(1+\frac{||\bm{\theta}||_2^2}{d\sigma_q^2})\right]+\ln\frac{3}{\delta}+\frac{1}{4}+\Theta(K\epsilon_\text{all})\right)
\end{aligned}
    \label{eq:pacgnn2}
\end{equation}
As $\mathcal{Q}\sim\mathcal{N}({\bm{\theta}},{\sigma}_{q}^2\bm{I}_d)$
, consider ${z}\sim\mathcal{N}({0},{\sigma}_{q}^2\bm{I}_d)$, we have $\frac{{z}}{{\sigma}_{q}}\sim\mathcal{N}({0},I_d)$ and $\mathbb{E}_{\theta\sim\mathcal{Q}}[L_{\text{train}}(\theta;A,X)] = \mathbb{E}_{z}[L_{\text{train}}(\theta+z;A,X)]$. Thus we have $\forall m>0$
\begin{equation}
    \begin{aligned}
    &\mathbb{E}_{\bm{\theta}\sim\mathcal{Q}}[L_{\text{train}}(\bm{\theta};\bm{A},\bm{X})]\\ &= \mathbb{E}_{\bm{z}}[L_{\text{train}}(\bm{\theta}+\bm{z};\bm{A},\bm{X})]\\
    & = \mathbb{E}_{\bm{z}}[L_{\text{train}}(\bm{\theta}+\bm{z};\bm{A},\bm{X})|\,||\frac{{\bm{z}}}{{\sigma}_{q}}||_2\leq m]\mathbb{P}(||\frac{{\bm{z}}}{{\sigma}_{q}}||_2\leq m)
    \\
    &+\mathbb{E}_{\bm{z}}[L_{\text{train}}(\bm{\theta}+\bm{z};\bm{A},\bm{X})|\,||\frac{{\bm{z}}}{{\sigma}_{q}}||_2> m]\mathbb{P}(||\frac{{\bm{z}}}{{\sigma}_{q}}||_2> m)
    )\\
    &\leq \max_{||\bm{\delta}||_2\leq m\sigma_q}[L_{\text{train}}(\bm{\theta}+\bm{\delta};\bm{A},\bm{X})]\mathbb{P}(||\frac{{\bm{z}}}{{\sigma}_{q}}||_2\leq m)+\mathbb{P}(||\frac{{\bm{z}}}{{\sigma}_{q}}||_2> m).\\
    &\leq \max_{||\bm{\delta}||_2\leq m\sigma_q}[L_{\text{train}}(\bm{\theta}+\bm{\delta};\bm{A},\bm{X})]+\mathbb{P}(||\frac{{\bm{z}}}{{\sigma}_{q}}||_2> m).
    \end{aligned}
\end{equation}
As $\frac{{\bm{z}}}{{\sigma}_{q}}\sim\mathcal{N}({0},I_d)$, by Chernoff bound of chi-squred distribution we have when $m\geq\sqrt{d}$,
\begin{equation}
    \begin{aligned}
     \mathbb{P}(||\frac{{\bm{z}}}{{\sigma}_{q}}||_2> m) \leq (\frac{m^2}{d}e^{1-\frac{m^2}{d}})^{d/2}
    \end{aligned}
\end{equation}
Thus  \begin{equation}
    \mathbb{E}_{\bm{\theta}\sim\mathcal{Q}}[L_{\text{train}}(\bm{\theta};\bm{A},\bm{X})]\leq \max_{||\bm{\delta}||_2\leq m\sigma_q}[L_{\text{train}}(\bm{\theta}+\bm{\delta};\bm{A},\bm{X})]+(\frac{m^2}{d}e^{1-\frac{m^2}{d}})^{d/2},
\end{equation}
combining it with Eq. \ref{eq:pacgnn2} and let $\sigma_q=\rho/m$ we have 
\begin{equation}
\begin{aligned}
L_{\text{all}}(\bm{\theta};\bm{A},\bm{X})\leq\max_{||\bm{\delta}||_2\leq \rho}[L_{\text{train}}(\bm{\theta}+\bm{\delta};\bm{A},\bm{X})]+(\frac{m^2}{d}e^{1-\frac{m^2}{d}})^{d/2}\\
+\frac{1}{\sqrt{N_0}}\left(\frac{1}{2}\left[1+d\log(1+\frac{m^2||\bm{\theta}||_2^2}{d\rho^2})\right]+\ln\frac{3}{\delta}+\frac{1}{4}+\Theta(K\epsilon_\text{all})\right).
\end{aligned}
\end{equation}
\end{proof}
\section{Proof of Theorem 2}
\label{sec:proofs}
\begin{proof} (Theorem \ref{th:invariant})
For ease of calculation we denote $L_\text{train}({\bm{\theta}};\bm{A},\bm{X})$ by $L({\bm{\theta}})$.
We need to show a) $\nabla_{{\bm{\theta}}}L({\bm{\theta}}+\nabla_{{\bm{\theta}}}L({\bm{\theta}}))|_{{\bm{\theta}}*} = 0$, and b) $\Delta_{{\bm{\theta}}}L({\bm{\theta}}+\nabla_{{\bm{\theta}}}L({\bm{\theta}}))|_{{\bm{\theta}}*} $ is positive definite \cite{wu2022fast,wu2022faster,wu2023decentralized}.

a) Since ${\bm{\theta}}^*$ is a local minimum of $L$, we have $\nabla_{{\bm{\theta}}}L({\bm{\theta}})|_{{\bm{\theta}}^*} = 0$, thus
\begin{equation}
\begin{aligned}
    \nabla_{{\bm{\theta}}}L({\bm{\theta}}+\nabla_{{\bm{\theta}}}L({\bm{\theta}}))|_{{\bm{\theta}}*} &= (I+\Delta_{{\bm{\theta}}}L({\bm{\theta}}^*))\nabla_{{\bm{\theta}}}L({\bm{\theta}})|_{{\bm{\theta}}^*+\nabla_{{\bm{\theta}}}L({\bm{\theta}})|_{{\bm{\theta}}^*}}\\
    &= (I+\Delta_{{\bm{\theta}}}L({\bm{\theta}}^*))\nabla_{{\bm{\theta}}}L({\bm{\theta}})|_{{\bm{\theta}}^*}  = 0
\end{aligned}
\end{equation}

b)\begin{equation}
\begin{aligned}
    \nabla_{{\bm{\theta}}}(\nabla_{{\bm{\theta}}}L({\bm{\theta}}+\nabla_{{\bm{\theta}}}L({\bm{\theta}})))|_{{\bm{\theta}}*} &= \nabla_{{\bm{\theta}}}[(I+\Delta_{{\bm{\theta}}}L({\bm{\theta}}))\nabla_{{\bm{\theta}}}L({\bm{\theta}}+\nabla_{{\bm{\theta}}}L({\bm{\theta}}))]|_{{\bm{\theta}}^*}\\
    &= \nabla_{{\bm{\theta}}}(I+\Delta_{{\bm{\theta}}}L({\bm{\theta}}))|_{{\bm{\theta}}^*}\nabla_{{\bm{\theta}}}L({\bm{\theta}})|_{{\bm{\theta}}^*+\nabla_{{\bm{\theta}}}L({\bm{\theta}})|_{{\bm{\theta}}^*}}\\
    &+(I+\Delta_{{\bm{\theta}}}L({\bm{\theta}}))|_{{\bm{\theta}}^*}\Delta_{{\bm{\theta}}}L({\bm{\theta}})|_{{\bm{\theta}}^*+\nabla_{{\bm{\theta}}}L({\bm{\theta}})|_{{\bm{\theta}}^*}}(I+\Delta_{{\bm{\theta}}}L({\bm{\theta}}))^T|_{{\bm{\theta}}^*}\\
    &=(I+\Delta_{{\bm{\theta}}}L({\bm{\theta}}))|_{{\bm{\theta}}^*}\Delta_{{\bm{\theta}}}L({\bm{\theta}})|_{{\bm{\theta}}^*}(I+\Delta_{{\bm{\theta}}}L({\bm{\theta}}))^T|_{{\bm{\theta}}^*} \\
\end{aligned}
\end{equation}
Because $(I+\Delta_{{\bm{\theta}}}L({\bm{\theta}}))|_{{\bm{\theta}}^*}$ and $\Delta_{{\bm{\theta}}}L({\bm{\theta}})|_{{\bm{\theta}}^*}$ are positive definite matrices, and $\nabla_{{\bm{\theta}}}(\nabla_{{\bm{\theta}}}L({\bm{\theta}}+\nabla_{{\bm{\theta}}}L({\bm{\theta}})))|_{{\bm{\theta}}*} = (I+\Delta_{{\bm{\theta}}}L({\bm{\theta}}))|_{{\bm{\theta}}^*}\Delta_{{\bm{\theta}}}L({\bm{\theta}})|_{{\bm{\theta}}^*}(I+\Delta_{{\bm{\theta}}}L({\bm{\theta}}))^T|_{{\bm{\theta}}^*}$ is symmetric, $\nabla_{{\bm{\theta}}}(\nabla_{{\bm{\theta}}}L({\bm{\theta}}+\nabla_{{\bm{\theta}}}L({\bm{\theta}})))|_{{\bm{\theta}}*}$ is positive definite. Thus ${\bm{\theta}}^*$ is also the local minimum of $L({\bm{\theta}}+\nabla_{{\bm{\theta}}}L({\bm{\theta}}))$.

\end{proof}

\section{Generalization Gap on i.i.d. Graph-level Tasks}
\label{sec:gen_gap_graph_level}
\begin{theorem}
(generalization bound) Assuming $L_{\textup{all}}(\bm{\theta};\bm{A},\bm{X})\leq\mathbb{E}_{\bm{z}\sim\mathcal{N}(\bm{0},\bm{\Sigma})}[L_{\textup{all}}(\bm{\theta}+\bm{z};\bm{A},\bm{X})]$, for any set of training nodes $\mathcal{V}_\textup{train}$ from $\mathcal{V}_\textup{all}$, $\forall m>\sqrt{d}$, with probability at least $1-\delta$, we have
\begin{equation}
\begin{aligned}
L_{\textup{all}}(\bm{\theta};\bm{A},\bm{X})&\leq\max_{||\bm{\delta}||_2\leq \rho}[L_{\textup{train}}(\bm{\theta}+\bm{\delta};\bm{A},\bm{X})]+(\frac{m^2}{d}e^{1-\frac{m^2}{d}})^{d/2}\\
&+\mathcal{O}\left(\sqrt\frac{d^{l-1}h\log(lh)\prod_{i=1}^{l}||\bm{\theta}_i||_2^2\sum_{i=1}^{l}(||\bm{\theta}_i||_F^2/||\bm{\theta}_i||_2^2)/\rho^2+\log(ml/\delta)}{m}\right)
\end{aligned}
\end{equation}
where $d$ is the number of parameters in the GCN, $l$ is the number of layers in GCN, $\bm{\theta}_i$ is the weight of i-th layer in GCN.
\end{theorem}
We use the assumption in \citet{foret2020sharpness}, $L_{\text{all}}(\bm{\theta};\bm{A},\bm{X})\leq\mathbb{E}_{\bm{z}\sim\mathcal{N}(\bm{0},\bm{\Sigma})}[L_{\text{all}}(\bm{\theta}+\bm{z};\bm{A},\bm{X})]$, which means that adding Gaussian perturbation should not decrease the test error.
\begin{proof}
\begin{lemma}(PAC-Bayes bound for GCN on node classification tasks \cite{liao2020pac})
For any $l>1$, $\rho > 0$, let $f \in \mathcal{H} : \mathcal{X} \times \mathcal{G} \to \mathbb{R}^K$ be an $l$ layer GCN with weights $\bm{W}_1,...,\bm{W}_l$. $h$ is the maximum hidden dimension, $\mathcal{P}$ and $\mathcal{Q}$ denote the prior and posterior distribution of weights. Then for any $\delta,\gamma > 0$, with probability at least $1-\delta$ over the choice of an i.i.d. size-$m$ training set $\mathcal{V}_\textup{train}\in\mathcal{V}_\textup{all}$, there exist a prior $\mathcal{P}$, such that
\begin{equation*}
    \mathbb{E}_{\bm{\theta}\sim\mathcal{Q}}[L_{\textup{all}}(\bm{\theta};\bm{A},\bm{X})]\leq  \mathbb{E}_{\bm{\theta}\sim\mathcal{Q}}[L_{\textup{train}}(\bm{\theta};\bm{A},\bm{X})]+
    \mathcal{O}\left(\sqrt\frac{d^{l-1}h\log(lh)\prod_{i=1}^{l}||\bm{W}_i||_2^2\sum_{i=1}^{l}(||\bm{W}_i||_F^2/||\bm{W}_i||_2^2)/\rho^2+\log(ml/\delta)}{m}\right)
\end{equation*}
\end{lemma}
Following the same discussion as in Sec. \ref{sec:genebound}, we have $L_{\text{all}}(\bm{\theta};\bm{A},\bm{X})\leq\mathbb{E}_{\bm{z}\sim\mathcal{N}(\bm{0},\bm{\Sigma})}[L_{\text{all}}(\bm{\theta}+\bm{z};\bm{A},\bm{X})]$ and 
\begin{equation}
    \mathbb{E}_{\bm{\theta}\sim\mathcal{Q}}[L_{\text{train}}(\bm{\theta};\bm{A},\bm{X})]\leq \max_{||\bm{\delta}||_2\leq \rho}[L_{\text{train}}(\bm{\theta}+\bm{\delta};\bm{A},\bm{X})]+(\frac{m^2}{d}e^{1-\frac{m^2}{d}})^{d/2},
\end{equation}
which yields
\begin{equation}
\begin{aligned}
L_{\text{all}}(\bm{\theta};\bm{A},\bm{X})&\leq\max_{||\bm{\delta}||_2\leq \rho}[L_{\text{train}}(\bm{\theta}+\bm{\delta};\bm{A},\bm{X})]+(\frac{m^2}{d}e^{1-\frac{m^2}{d}})^{d/2}\\
&+\mathcal{O}\left(\sqrt\frac{d^{l-1}h\log(lh)\prod_{i=1}^{l}||\bm{\theta}_i||_2^2\sum_{i=1}^{l}(||\bm{\theta}_i||_F^2/||\bm{\theta}_i||_2^2)/\rho^2+\log(ml/\delta)}{m}\right)
\end{aligned}
\end{equation}
\end{proof}
\begin{table}[h]
\vspace{-5pt}
\caption{Performance of WT-AWP on graph classification.}
\vspace{-8pt}
\label{tab:graphclass}
\centering
\scalebox{0.85}{
\begin{tabular}{cccc}
\hline\hline
{ } & Proteins              & IMDB-Binary           & IMDB-Multi            \\ \hline
GCN                     & 75.05 ± 1.40          & 72.40 ± 2.73          & 55.53 ± 1.33          \\
GCN+WT-AWP              & \textbf{76.48 ± 0.49} & \textbf{75.80 ± 1.17} & \textbf{57.26 ± 0.63} \\ \hline
\end{tabular}
}
\vspace{-15pt}
\end{table}
\subsection{Graph Classification}
\label{sec:graph_classification}

Finally, we conduct experiments on graph classification tasks with three benchmark datasets: Protein, IMDB-Binary and IMDB-Multi. Detailed description is in Sec. \ref{sec:datagraphc}. Table \ref{tab:graphclass} shows the experimental results. Generally, WT-AWP improves the accuracy with a large margin. Besides, the variance of the accuracy of GCN+WT-AWP across different random seeds is significantly smaller than the vanilla GCN, which indicates that WT-AWP is also more stable.
Note, we do not claim that our models are state of the art, but rather that WT-AWP provides consistent improvements. Besides, graph classification tasks are i.i.d. and our theoretical results may not hold for them, we put an analysis of these i.i.d. graph level tasks in Appendix C, which also demonstrates the effectiveness of AWP methods on graph level tasks.
\section{Vanishing-gradient Issue of AWP on MLP}\label{sec:vanishmlp}
In this section we show that the vanishing-gradient issue also happens in multi-layer perceptrons.
Consider an MLP $\hat{y}=\sigma(\bm{W}_n(...(\bm{W}_1\bm{X})))$ with a softmax activation at the output layer. The perturbed model is $\hat{y}=\sigma(((\bm{W}_n+{\bm{\delta}}_n)(...((\bm{W}_1+{\bm{\delta}}_1)\bm{X})))$. Since the norm of each perturbation ${\bm{\delta}}_i$ could be as large as $\rho||\bm{W}_i||_2$, in the worst case the norm of each layer is $(\rho+1)||\bm{W}_i||_2$, and thus the model will have exploding logit values when $\rho$ is large. After feeding large logits into the softmax layer, the output will approximate a one-hot encoded vector, because the difference between the entries of the logits will also be large. The gradient will be close to $0$ and the weights will not be updated.\looseness=-1

\begin{figure}[H]
	\subfigure[Natural model]{
		\begin{minipage}[t]{.18\linewidth}
			\centering
			\includegraphics[height=2.5cm]{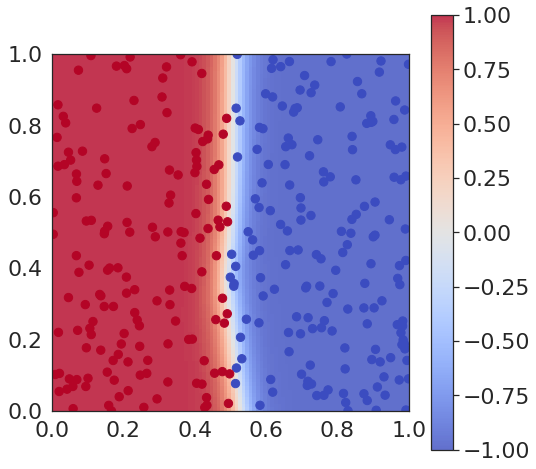}	
		\end{minipage}
	}
	\subfigure[AWP $\rho = 0.1$]{
		\begin{minipage}[t]{.18\linewidth}
			\centering
			\includegraphics[height=2.5cm]{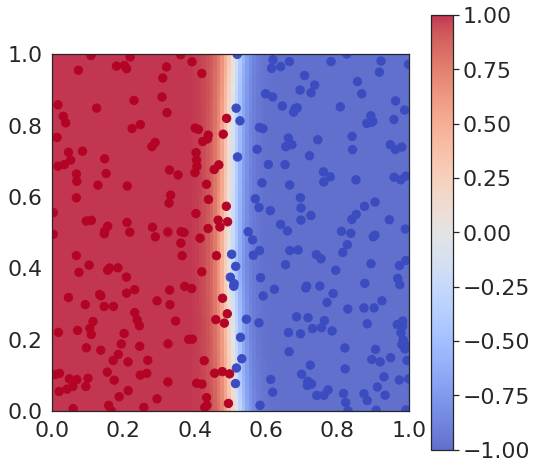}	
		\end{minipage}
	}
	\subfigure[AWP $\rho = 0.2$]{
		\begin{minipage}[t]{.18\linewidth}
			\centering
			\includegraphics[height=2.5cm]{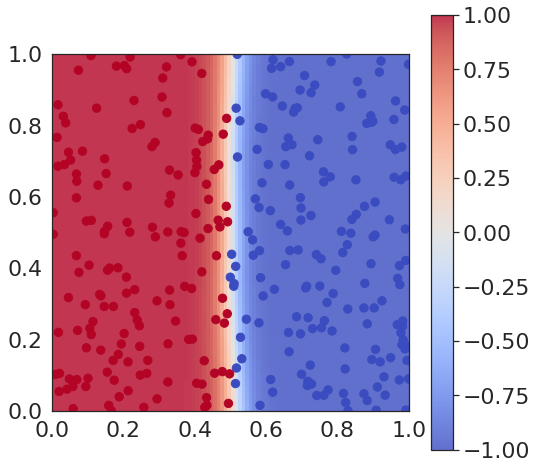}	
			\label{fig:ghe}
		\end{minipage}
	}
	\subfigure[AWP $\rho = 0.23$]{
		\begin{minipage}[t]{.18\linewidth}
			\centering
			\includegraphics[height=2.5cm]{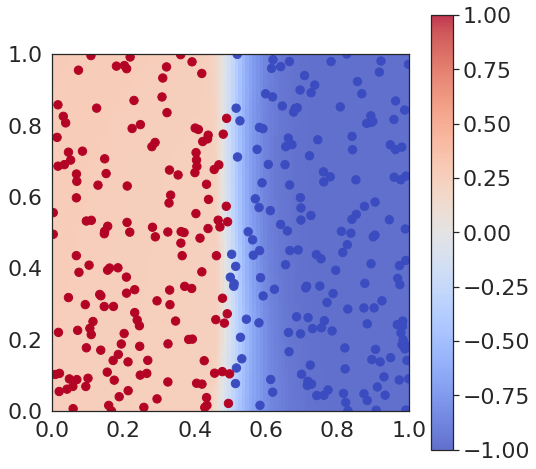}	
			\label{fig:linear1}
		\end{minipage}
	}
	\subfigure[AWP $\rho = 0.25$]{
		\begin{minipage}[t]{.18\linewidth}
			\centering
			\includegraphics[height=2.5cm]{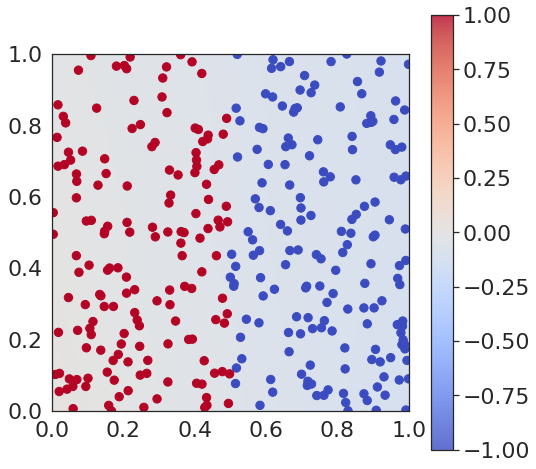}	
			\label{fig:linear}
		\end{minipage}
	}
	\caption{Comparing AWP models on a linearly separable dataset with different $\rho$ values.}
	\label{fig:aa}
\end{figure}

To verify our conclusion we train a 3-layer linear network with $\bm{W}_1\in\mathbb{R}^{2\times100}, \bm{W}_2\in\mathbb{R}^{100\times100},  \bm{W}_3\in\mathbb{R}^{100\times2}$ on a linearly separable dataset, the number of training epochs is 2000. 
In Figure \ref{fig:aa} we show the trained classifiers with different $\rho$. We find that models with AWP are crushed quickly when $\rho$ increased from $0.2$ to $0.25$. When $\rho = 0.25$, the value of loss function remains unchanged during training and the prediction score is around $0$, which indicates the weights are almost not updated during training. So with the AWP objective, we cannot select a large $\rho$. 

\begin{figure}[t]
	\subfigure[Natural model]{
		\begin{minipage}[t]{.23\linewidth}
			\centering
			\includegraphics[height=3cm]{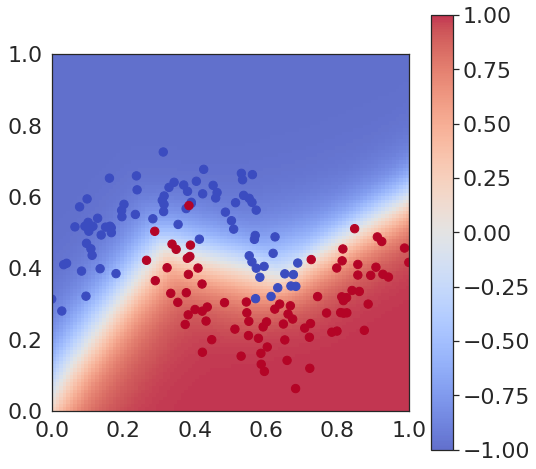}	
			\label{fig:aaaa}
		\end{minipage}
	}
	\subfigure[AWP]{
		\begin{minipage}[t]{.23\linewidth}
			\centering
			\includegraphics[height=3cm]{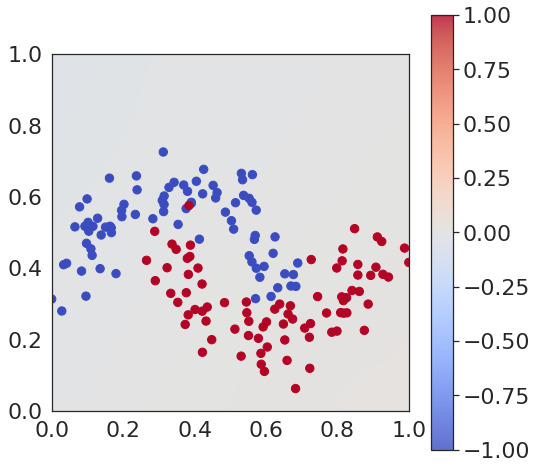}	
			\label{fig:aaab}
		\end{minipage}
	}
	\subfigure[T-AWP]{
		\begin{minipage}[t]{.23\linewidth}
			\centering
			\includegraphics[height=3cm]{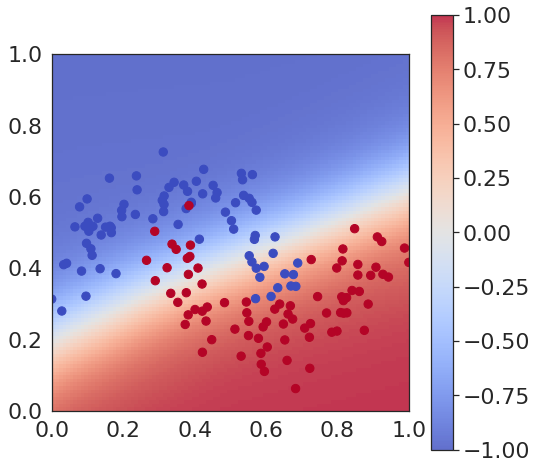}	
			\label{fig:aaac}
		\end{minipage}
	}
	\subfigure[W-AWP, $\lambda=0.9$]{
		\begin{minipage}[t]{.23\linewidth}
			\centering
			\includegraphics[height=3cm]{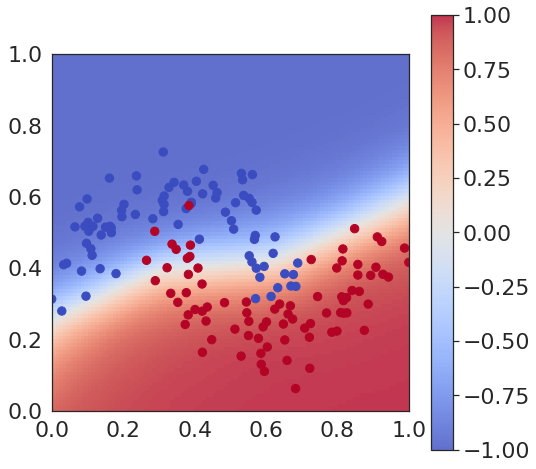}	
			\label{fig:aaad}
		\end{minipage}
	}
	\caption{Model comparison on the two moons dataset, $\rho = 0.4$}
	\label{fig:aaa}
\end{figure}

We repeat the experiment using the same 3-layer linear network on a 2d moons dataset and perform the weight perturbation only on the first two layers. 
Figure \ref{fig:aaa} Illustrates the trained models with $\rho = 0.4$. In  Figure \ref{fig:aaab} we can see the model suffers from vanishing gradients, while the model with truncated AWP works well (Figure \ref{fig:aaac}). Besides, comparing to the overfitted natural model (Figure \ref{fig:aaaa}), the truncated AWP model has a smoother decision boundary. We also remove the weight perturbation on the middle or the first layer and train the model correspondingly, the results are similar to Figure \ref{fig:aaac}. As Figure \ref{fig:aaad} shows, the model with the weighted AWP objective is able to learn the representation of the input data, and the decision boundary is also smoother than the nature model in Figure \ref{fig:aaaa}.

\subsection{Norm of gradient during training with large $\rho$}
\begin{figure*}[t]
		\begin{minipage}[t]{.32\linewidth}
			\centering
			\includegraphics[height=3.5cm]{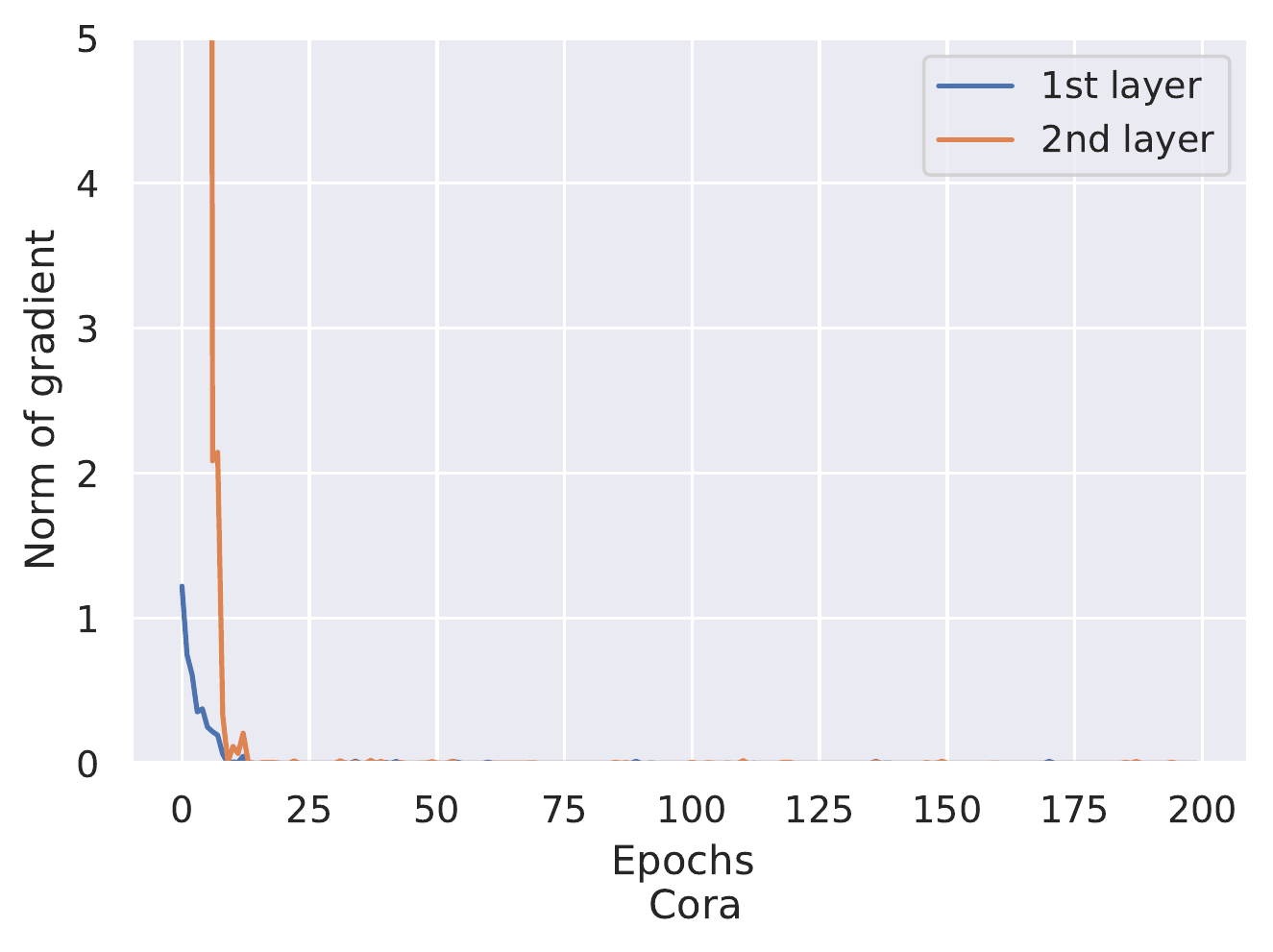}
			\vspace{-6pt}
		\end{minipage}
		\begin{minipage}[t]{.32\linewidth}
			\centering
			\includegraphics[height=3.5cm]{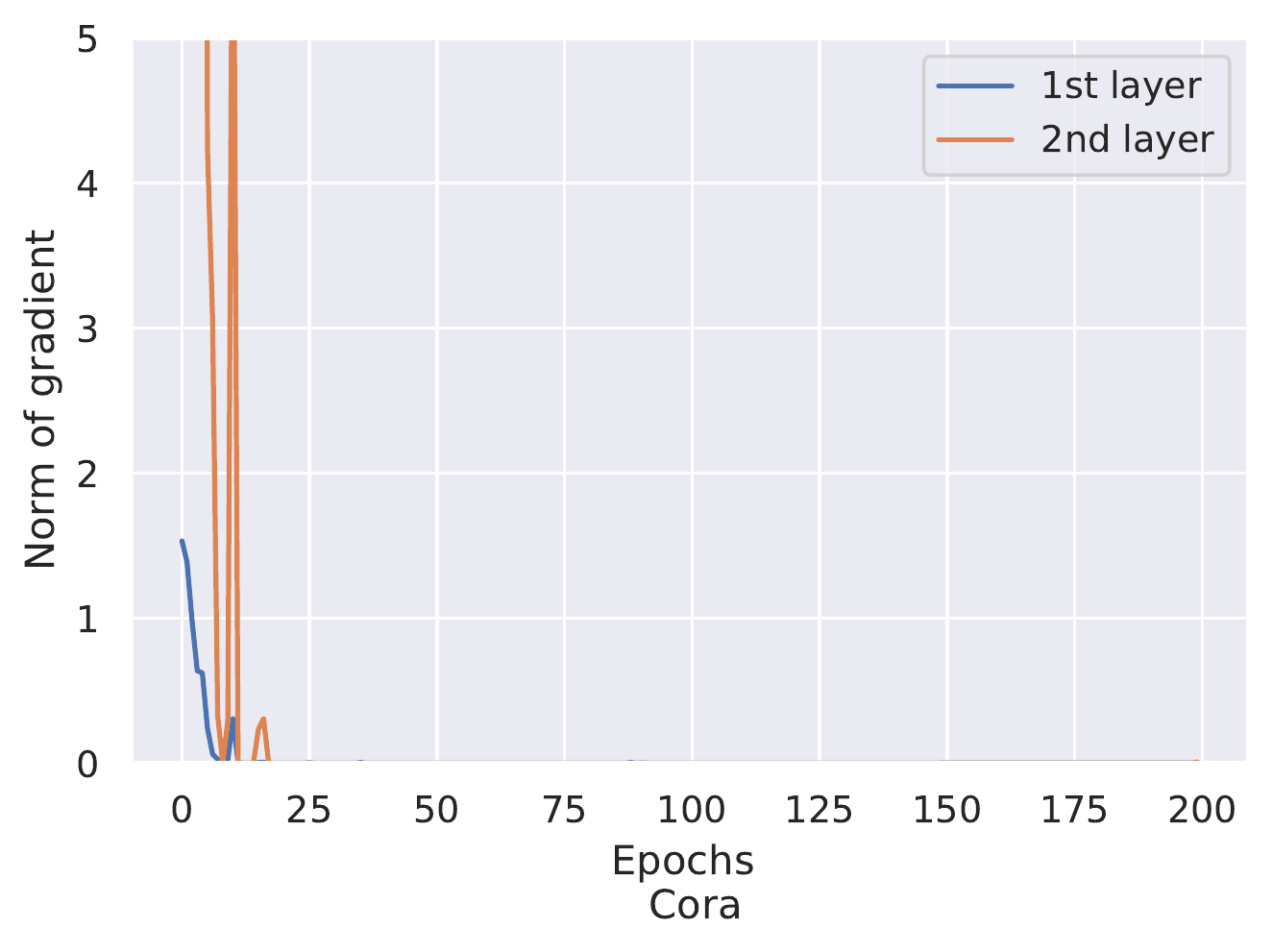}	
			\vspace{-6pt}
		\end{minipage}
		\begin{minipage}[t]{.32\linewidth}
			\centering
			\includegraphics[height=3.5cm]{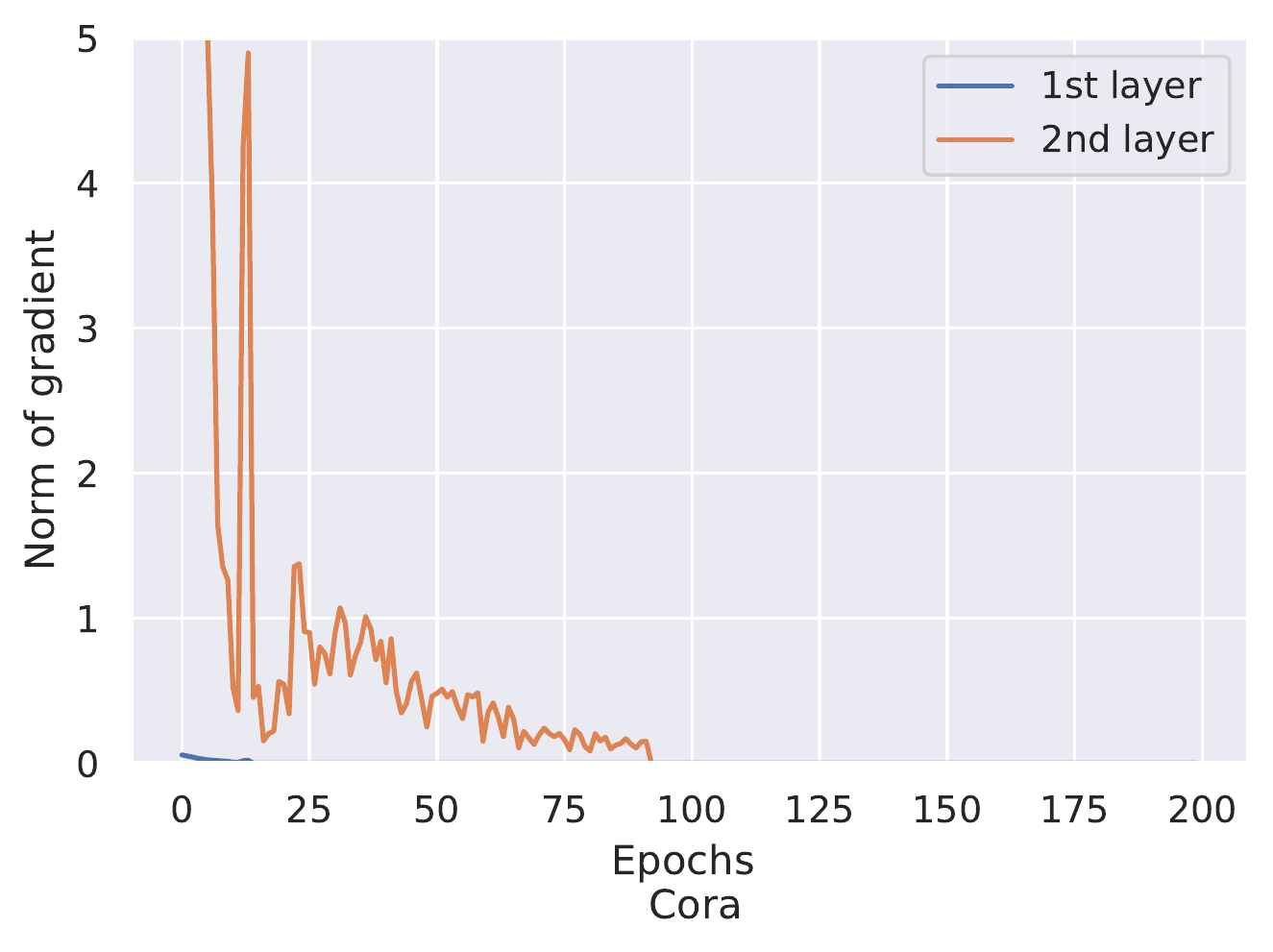}	
			\vspace{-6pt}
		\end{minipage}
		
		\caption{Norm of gradient during training with large $\rho$, the settings are the same as in \ref{sec:cleanaccuracy}.}
		\label{fig:gdnorm_all}
		\vspace{-12pt}
	\end{figure*}

In order to show the vanish gradient phenomenon during training, we conduct experiments on GCN+AWP model with $\rho=20$ on Cora, Citeseer, and Polblogs datasets. The detailed settings are the same as in \ref{sec:cleanaccuracy}. \ref{fig:gdnorm_all} illustrates the results. From the plots we can observe the vanish gradient issues on Cora and Citeseer from the 20th epoch, and on Polblogs from 80th epoch, which is consistency with our above analysis.

\subsection{Description of Algorithm \ref{alg:2}}\label{sec:despalg}
In the WT-AWP algorithm, we apply a numerical optimizer such as Adam to the WT-AWP objective 
$$ L_\text{WT-AWP}(\bm{\theta}) = [\lambda L_\text{train}({{\bm{\theta}}}+[\hat{{\bm{\delta}}}^\text{(awp)*}({{\bm{\theta}}}^\text{(awp)}),0];{\bm{A}},{\bm{X}})+(1-\lambda)L_\text{train}({{\bm{\theta}}};{\bm{A}},{\bm{X}})]$$
Since in our empirical experiments the GNNs always have a 2-layer structure, we assign $\bm{\theta}^\text{awp}=\bm{W}_1$ (the first layer) and $\bm{\theta}^\text{normal}=\bm{W}_2$ (the last layer).
Then the perturbation $\hat{{\bm{\delta}}}^\text{(awp)*}$ (first layer) is computed via Eq. \ref{eq:est}. Next the gradient $g$ of $L_\text{WT-AWP}$ is calculated as in Eq. \ref{eq:grad_awp},
$$\bm{g} = \lambda\nabla_{{\bm{\theta}}} L_\text{train}({{\bm{\theta}}};{\bm{A}},{\bm{X}})|_{{{\bm{\theta}}}_{t-1}+[\hat{{{\bm{\delta}}}}^*({{\bm{\theta}}}_{t-1}^{(\text{awp})}),0]}+(1-\lambda)\nabla_{{\bm{\theta}}} L_\text{train}({{\bm{\theta}}};{\bm{A}},{\bm{X}})|_{{{\theta}}_{t-1}}$$.\\
Finally we update the weight via ${{\bm{\theta}}}_t = {{\bm{\theta}}}_{t-1}-\alpha \bm{g}$.
	
\section{Additional Experiments}
\subsection{Learning Curves and Generalization Gap During Training}
\begin{figure}[]
    
	\subfigure[Learning curves]{
		\begin{minipage}[t]{.5\linewidth}
			\centering
			\includegraphics[height=5cm]{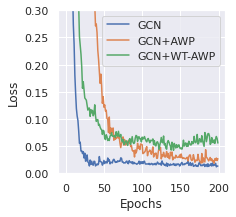}	
			\label{fig:learncurve}
		\end{minipage}
	}
	\subfigure[Generalization gap]{
		\begin{minipage}[t]{.5\linewidth}
			\centering
			\includegraphics[height=5cm]{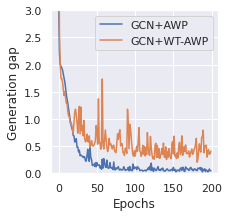}	
			\label{fig:gap}
		\end{minipage}
	}
	\caption{Learning Curves and Generalization Gap During Training.}
	\label{fig:lcagg}
\end{figure}

In this part, we train GCN, GCN+AWP ($\rho=0.1$) and GCN+WT-AWP ($\lambda = 0.5, \rho = 0.5$) models with the same random initialization and compare their learning curves, and generalization gap. The accuracy is 0.8355 for GCN, 0.8421 for GCN+AWP, and 0.8551 for GCN+WT-AWP. 
 Figure \ref{fig:learncurve} illustrates the learning curve of vanilla loss $L_\text{train}({\bm{\theta}};\bm{A},\bm{X})$ during training. The loss of all three models converges well. The final value of GCN+WT-AWP is larger than the rest two models. We believe it is because GCN+WT-AWP finds a different (flatter) local minimum.
Figure \ref{fig:gap} shows the generalization gap during training. Because we use a large perturbation bound $\rho$ in GCN+WT-AWP, its generalization gap fluctuates more and decrease slower compared to the gap of GCN+AWP. The fluctuation is due to the exploding logit problem in AWP with a large $\rho$ value. When it happens, the regular loss included in WT-AWP can minimize (but not completely eliminate) its influence. 
Despite the fluctuation, the generalization gap of GCN+WT-AWP decreases with time as well.
\subsection{Robust Accuracy with a Poisoning DICE Attack}\label{sec:poisoningdice}
We conduct additional experiments on poisoning the graph with DICE attacks. The general model settings are the same as Sec. \ref{sec:poisoning}. The WT-AWP hyperparameters $(\lambda,\rho)$ are shown in Table \ref{tab:dicepoisonpar}. Table \ref{tab:dicepoison} illustrates the experimental results. The models that achieve best performance on a given dataset are all based on WT-AWP. Besides, WT-AWP also consistently boost the performance of the baselines.
\begin{table}[]
\centering
\caption{Robust accuracy with 5\% poisoning DICE attacks. We report the average and the standard deviation across 200 experiments per model (20 random splits $\times$ 10 random initializations).}
\vspace{3pt}

\begin{tabular}{ccccccc}
\hline
\hline
              & \multicolumn{3}{c}{Natural Acc.}                                      & \multicolumn{3}{c}{Acc. with 5\% DICE attack}                                                               \\ \hline
Approachs     & Cora                  & Citeseer              & Polblogs              & Cora                                                        & Citeseer              & Polblogs              \\ \hline
GCN           & 83.73 ± 0.71          & 73.03 ± 1.19          & 95.06 ± 0.68          & 82.60 ± 0.76                                                & 71.89 ± 1.17          & 90.13 ± 0.82          \\
+WT-AWP        & \textbf{84.66 ± 0.53} & 74.01 ± 1.11          & \textbf{95.20 ± 0.61} & \textbf{83.87 ± 0.62}                                       & 73.68 ±1.06           & 90.31 ± 0.79          \\ \hline
GCNJaccard    & 82.42 ± 0.73          & 73.09 ± 1.20          & N/A                   & 81.55 ± 0.86                                                & 72.22 ± 1.22          & N/A                   \\
+WT-AWP & 83.55 ± 0.60          & 74.10 ± 1.04          & N/A                   & 82.86 ± 0.73                                                & \textbf{73.95 ± 1.04} & N/A                   \\ \hline
SimPGCN       & 82.99 ± 0.68          & 74.05 ± 1.28          & 94.67 ± 0.95          & 82.11 ± 0.70                                                & 73.53 ± 1.23          & 89.57 ± 1.06          \\
+WT-AWP    & 83.37 ± 0.74          & \textbf{74.26 ± 1.09} & 94.85 ± 0.91          & 83.30 ± 0.73                                                & 73.89 ± 1.08          & 90.13 ± 1.03          \\ \hline
GCNSVD        & 77.63 ± 0.63          & 68.57 ± 1.54          & 94.08 ± 0.59          &  76.25 ± 0.91 & 67.27 ± 1.67          & 90.80 ± 0.88          \\
+WT-AWP     & 79.05 ± 0.58          & 71.12 ± 1.42          & 94.13 ± 0.59          & 77.51 ± 0.77 & 70.30 ± 1.22          & \textbf{91.11 ± 0.76} \\ \hline
RGCN          & 83.29 ± 0.63          & 71.69 ± 1.35          & 95.15 ± 0.46          & 82.02 ± 0.73                                                & 70.18 ± 1.38          & 90.03 ± 0.67          \\ \hline
\end{tabular}

\label{tab:dicepoison}
\end{table}

\begin{table}[H]
\centering
\caption{Hyperparameters of WT-AWP for poisoning DICE attacks}
\vspace{3pt}
\label{tab:dicepoisonpar}

\begin{tabular}{llll}
\hline\hline
$(\lambda,\rho)$                                                  & Cora       & Citeseer   & Polblogs                                                                      \\ \hline
\begin{tabular}[c]{@{}l@{}}GCN\\ GCNJaccard\\ GCNSVD\end{tabular} & (0.5, 0.5) & (0.7, 2)   & (0.3, 1)                                                                    \\ \hline
SimPGCN                                                           & (0.1, 0.5) & (0.5, 0.1) & (0.5, 1)  \\ \hline
\end{tabular}

\end{table}
 
 
\subsection{Certified Robustness on the Citeseer Dataset}\label{sec:certcite}
\begin{figure}[]
	\subfigure[Node feature perturbations]{
		\begin{minipage}[t]{.5\linewidth}
			\centering
			\includegraphics[height=4.5cm]{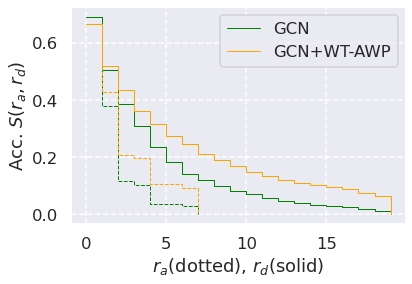}	
		\end{minipage}
	}
	\subfigure[Graph structure perturbations]{
		\begin{minipage}[t]{.5\linewidth}
			\centering
			\includegraphics[height=4.5cm]{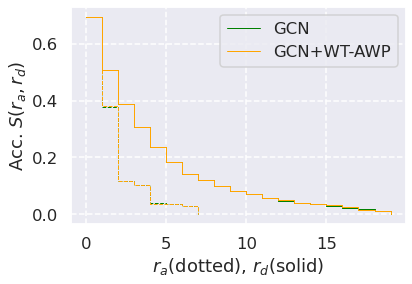}	
		\end{minipage}
	}
	\caption{Certified adversarial robustness on the Citeseer dataset.}
	\label{fig:certcite}
\end{figure}

We measure the certified robustness of GCN and GCN+WT-AWP with randomized smoothing \cite{bojchevski2020efficient} on the Citeseer dataset.  We use $\lambda = 0.5, \rho=1$ as the hyperparameters for WT-AWP models.
We plot the certified accuracy $S(r_a,r_d)$ w.r.t. $r_a$ and $r_d$. As seen in Figure \ref{fig:certcite}, comparing with the vanilla GCN, WT-AWP significantly increases the certified accuracy for perturbations to the node features for all radii, while having comparable performance for certification of the graph structure.

\subsection{Gradient Norm During Training}
In this experiment we train a vanilla GCN, GCN+AWP with $\rho=0.1$, GCN+WT-AWP with $\lambda = 0.5,\rho=1$ on Cora and plot the relative gradient norm $||\nabla\bm{\theta}||_2/||\bm{\theta}||_2$ during training. Both AWP and WT-AWP have small relative gradient norm compared to GCN when epoch is larger than 100.

\begin{figure}[h]
		
	\centering
	\includegraphics[height=5cm]{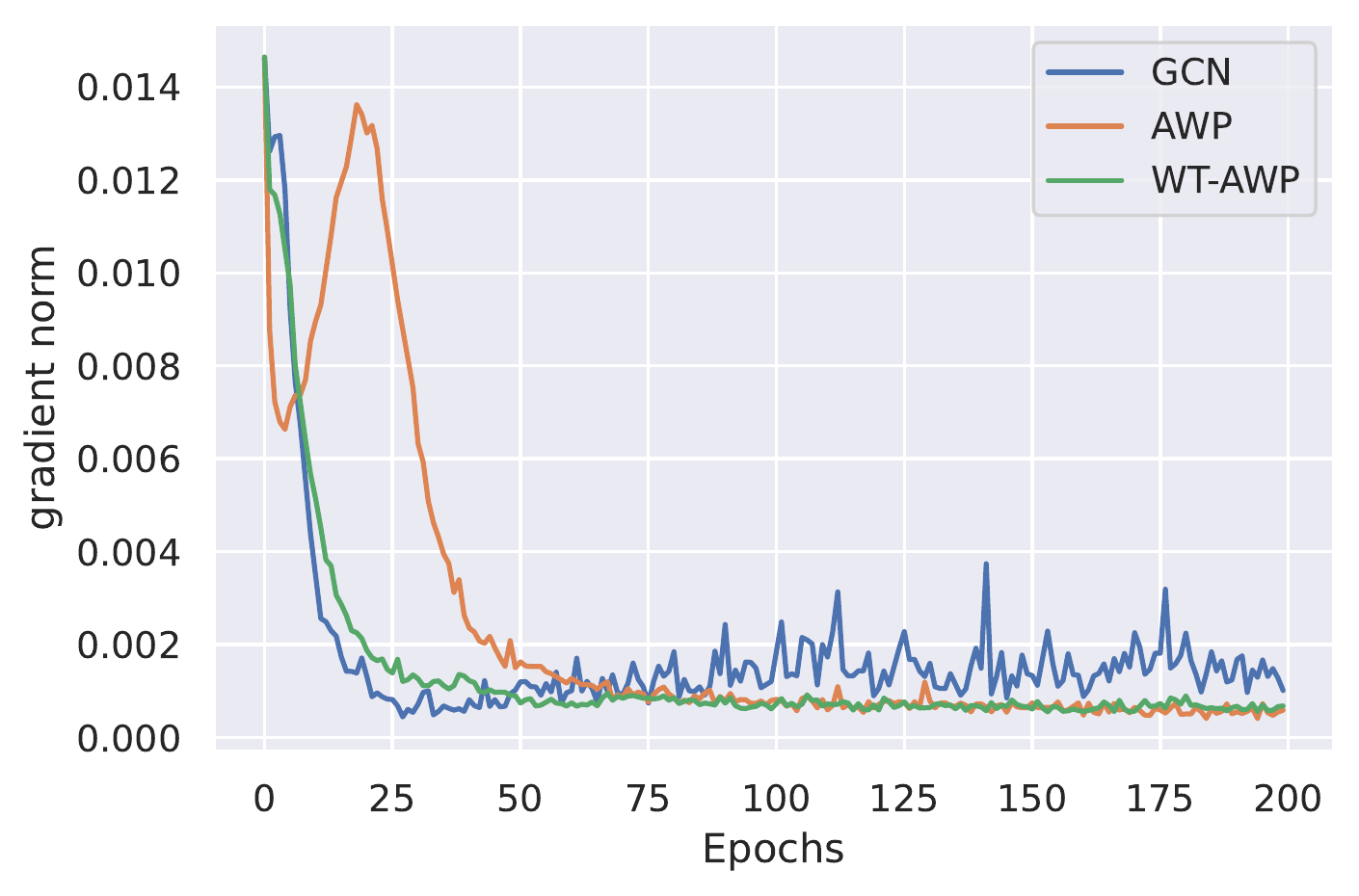}	

\caption{Norm of Gradient during training.}
\label{fig:ngrad}
\end{figure}
\subsection{Ablation Study of The Perturbed Layer}
Since in all experiment above we only perturb the first layer of GNN with WT-AWP, we provide experimental results corresponds to perturb only the second layer with WT-AWP. The backbone is GCN and the benchmark is Cora. As Table \ref{tab:ablationlast} shows, skipping the first layer in WT-AWP methods have worse performance than skipping the last layer.

\begin{table}[h]
\centering
\vspace{-6pt}
\caption{Ablation study with $\lambda$ and $\rho$ on WT-AWP, where we only use AWP on the last layer. The backbone model is GCN and the benchmark is Cora.}
\vspace{1pt}
\label{tab:ablationlast}
\scalebox{1}{
\begin{tabular}{c|cccccc}
\hline\hline
WT-AWP (last layer) & $\rho = 0.05$ & $\rho = 0.1$ & $\rho = 0.5$ & $\rho = 1$   & $\rho = 2.5$ & $\rho = 5$   \\ \hline
$\lambda = 0.1$             & 84.09 ± 0.62  & 84.13 ± 0.60 & 84.18 ± 0.60 & 83.87 ± 0.85 & 81.40 ± 1.92 & 65.34 ± 6.41 \\
$\lambda = 0.3$             & 84.14 ± 0.58  & 84.12 ± 0.64 & 84.14 ± 0.69 & 82.30 ± 1.47 & 33.50 ± 1.48 & 29.18 ± 0.00 \\
$\lambda = 0.5$             & 84.13 ± 0.60  & 84.10 ± 0.63 & 84.08 ± 0.77 & 78.00 ± 3.16 & 29.18 ± 0.00 & 29.18 ± 0.00 \\
$\lambda= 0.7$              & 84.12 ± 0.62  & 84.14 ± 0.64 & 83.74 ± 0.84 & 67.37 ± 5.01 & 29.18 ± 0.00 & 29.18 ± 0.00 \\
$\lambda = 1.0$             & 84.20 ± 0.62  & 84.19 ± 0.65 & 82.80 ± 1.04 & 29.18 ± 0.02 & 29.18 ± 0.00 & 29.18 ± 0.00 \\ \hline
\end{tabular}
}
\vspace{-12pt}
\end{table}

 \section{Experimental Details}

\subsection{Description of Baseline Models}\label{sec:baselinemodel}
We aim to evaluate the impact of our WT-AWP on natural and robust node classification tasks, thus we utilize the well-known graph neural networks and graph defense methods as baseline. We first train the baseline models and compare their performance with the baseline models trained with WT-AWP objective (if applicable). The baseline GNN models include:
\begin{itemize}

    \item \textbf{GCN} \cite{kipf2016semi}: is one of the most representative graph convolution neural networks. Currently it can still achieve SOTA on different graph learning tasks.
    \item \textbf{GAT} \cite{velivckovic2017graph}: utilizes multi-head attention mechanism to learn different weights for each node and its neighbor node without requiring the spectral decomposition. 

    \item \textbf{PPNP} \cite{klicpera2018predict}: improves the GCN propagation scheme based on the personalized Pagerank. This approach generates predictions from each node’s own features and propagates these predictions using an adaptation of personalized PageRank. 

    \item \textbf{RGCN} \cite{zhu2019robust}: applies the Gaussian distribution to model the node representations. This structure is expected to absorb effects of adversarial attacks. It also  penalizes nodes with large variance with an attention mechanism. Notice, the WT-AWP cannot be applied to RGCN, as the weights of RGCN are modeled by distributions. We can regard RGCN as another model inspired by PAC-Bayes theorem, as it models the objective $\mathbb{E}_{\bm{\delta}\sim\mathcal{N}(0,\sigma\bm{I})}[L_\text{train}({\bm{\theta}}+{\bm{\delta}};\bm{A},\bm{X})]$, which also bounded $L_\text{all}({\bm{\theta}};\bm{A},\bm{X})$ according to PAC-Bayes theorem.
    
    \item \textbf{GCNJaccard} \cite{wu2019adversarial}: is a graph defense method based on GCN. It pre-processes the graph by deleting edges, which connect nodes with a small Jaccard similarity of features, because attackers prefer connecting nodes with dissimilar features. This method only works on graph with node features. For example it cannot work on Polblogs because the node features are unavailable.
    
    \item \textbf{GCNSVD} \cite{entezari2020all}: is a graph defense method based on GCN, which focuses on defending nettack \cite{zugner2018adversarial}.
 Since nettack is a high-rank attack, GCN-SVD pre-processes the perturbed graph with its low-rank approximation. It is straightforward to extend it to non-targeted and random attacks.
 
    \item \textbf{SimpleGCN} \cite{jin2021node}: utilizes similarity preserving aggregation to integrate the graph structure and the node features, and employs self-supervised learning to capture the similarity between node features. Notice SimpleGCN is not specifically designed for graph defense, and we find it also has good performance under the poisoning attacks, thus we add this method as another graph defense baseline.
\end{itemize}

\subsection{Description of Graph Attack Methods}\label{sec:graphattacks}
Generally speaking, there are two types of the adversarial attacks on node classification tasks: test-time attack (evasion) and train-time attack (poisoning). In both types of attacks we first generate a perturbed adjacency matrix based on a victim model, and then in evasion attacks we test it directly on the victim model, and in poisoning attacks we train a new model with the perturbed adjacency matrix. For generating the adversarial perturbations, we apply three methods:
\begin{itemize}
\vspace*{-3pt}
\item \textbf{DICE} \cite{waniek2018hiding}: is a baseline attack method (delete internally, connect externally). In each perturbation, we randomly choose whether to insert or remove an edge. Edges are only removed between nodes from the same classes, and only inserted between nodes from different classes.
\item \textbf{PGD} \cite{xu2019topology}: calculates the gradient of the adjacency matrix, and the gradient serves as a probabilistic vector, then a random sampling is applied for generating a near-optimal binary perturbation based on this vector.
\item \textbf{Metattack} \cite{zugner2019adversarial}: was proposed to generate poisoning attacks based on meta-learning. It has an approximate version A-Metattack. In our experiments, we apply the original Metattack. 
\end{itemize}
\subsection{Datasets Statistics}
Cora and Citeseer \cite{sen2008collective} are citation datasets commonly used for evaluating GNNs. Polblogs \cite{adamic2005political} is another common benchmark dataset where each node is a political blog. In Table \ref{tab:dataset} we provide the statistics for each graph. We preprocess the graph and only use the largest connected component.
\label{sec:data_and_setup}
\begin{table}[H]
\centering
\caption{Dataset Statistics}
\vspace{0.1cm}
\begin{tabular}{c|ccc}
\hline
\hline
\textbf{Datasets} & \textbf{Cora} & \textbf{Citeseer} & \textbf{Polblogs} \\ \hline
\#Nodes           & 2708          & 3327              & 1222              \\
\#Edges           & 5429          & 4732              & 16714             \\
\#Features        & 1433          & 3703              & N/A               \\
\#Classes         & 7             & 6                 & 2                 \\ \hline
\end{tabular}
\label{tab:dataset}
\end{table}

\subsection{Training Setup}
\label{sec:training setup}
\textbf{Optimization hyperparameters.} We use the Adam optimizer with a learning rate 0.01 and weight decay of 0.0005. All models are trained for 200 epochs with no weight scheduling. We add a dropout layer with rate $p=0.5$ after each GNN layer during training. We apply no early stopping and the optimal model is selected with its performance on the validation set. The test set is never touched during training.

\noindent\textbf{Train/val/test split.} The evaluation procedures of GNNs on node classification tasks have suffered overfitting bias from using a single train-test split. \cite{shchur2018pitfalls} showed that different splits could significantly affect the performance and ranking of models. In all our experiments on node classification tasks, we apply the split setting in \cite{zugner2019adversarial}, which utilizes 10\% samples for training, 10\% samples for validating, and 80\% samples for testing. We generate 20 random splits and for each split we train 10 models with different random initialization. We report the mean and standard deviation of the accuracy of the 200 random models in our results.

\subsection{Settings of the Average of Gradient Norm}\label{sec:setagn}
For the results in Sec. \ref{sec:average_gradient} we generate noise $\bm{z_A,z_X}$ from Gaussian distribution $\mathcal{N}(\bm{A},\sigma^2\bm{I})$ and $\mathcal{N}(\bm{X},\sigma^2\bm{I})$, then calculate the $l_2$ norm of the loss gradient $||\nabla_{\bm{A}}L_\text{train}(\bm{\theta}; \bm{A}, \bm{X})|_{\bm{A=z_A}}||_2$ and $||\nabla_{\bm{X}}L_\text{train}(\bm{\theta}; \bm{A}, \bm{X})|_{\bm{A=z_A}}||_2$. In our experiments we choose $\sigma = 0.0005$, because we expect the perturbed input to be close to the clean input.

\subsection{Settings of Visualization of Loss Landscape} \label{sec:setvll}

For the results in Sec. \ref{sec:loss_landspace}, we generate a random direction $\bm{u}$ from a Gaussian distribution and perform $l_2$ normalization, it is equal to randomly selecting a direction on the $l_2$ unit ball. As seen in Figure \ref{fig:learncurve}, there is a large gap between the final loss value of WT-AWP and vanilla GCN, we have to parallel move the loss landscape of WT-AWP and GCN to the same level for making comparison. The experiments are performed on Cora, similar results also hold for other datasets.

\subsection{Hyperparameters $(\lambda,\rho)$ for Poisoning Attacks}\label{sec:hyppgdmeta}

\begin{table}[h]
\centering
\caption{Hyperparameters of WT-AWP for poisoning PGD attack and Metattack of Sec. \ref{sec:poisoning}}
\label{tab:papoi}
\vspace{3pt}
\begin{tabular}{llll}
\hline\hline
$(\lambda,\rho)$                                                  & Cora       & Citeseer   & Polblogs                                                                      \\ \hline
\begin{tabular}[c]{@{}l@{}}GCN\\ GCNJaccard\\ GCNSVD\end{tabular} & (0.7, 0.5) & (0.7, 2)   & (0.5, 0.5)                                                                    \\ \hline
SimPGCN                                                           & (0.3, 0.5) & (0.5, 0.1) & \begin{tabular}[c]{@{}l@{}}(0.3, 2)  Metattack\\ (0.5, 0.5)  PGD\end{tabular} \\ \hline
\end{tabular}
\end{table}

\subsection{Randomized Smoothing}
\label{sec:rand_smooth_details}
Although suffering from curse of dimensionality \cite{wu2021completing}, randomized smoothing is still one of the state-of-the-art certified robustness approches in many fields \cite{wu2022retrievalguard,cohen2019certified}.
Following \citet{bojchevski2020efficient}, we create smoothed versions of our GNN models by randomly perturbing the adjacency matrix (or the node features) and predicting the majority vote for the randomly-perturbed samples. 
We denote with $p_a$ the probability of flipping an entry from $0$ to $1$, i.e. adding an edge or a feature, and with $p_d$ the probability of flipping an entry from $1$ to $0$, \emph{i.e.} deleting an edge or a feature. In all experiments, for the certification of node features we generate random perturbations with $p_a = 0.01, p_d = 0.6$, and for perturbing the adjacency matrix we use $p_a = 0.001, p_d = 0.4$.
%
%
We consider the prediction of the smoothed GNN for a given node correct if and only if it is correct and certifiably robust. This means the prediction of the node does not change for any perturbation within the radius (i.e. for any $r_d$ deletions or $r_a$ additions). 

\subsection{Graph Classification}\label{sec:datagraphc}
\begin{table}[H]

\centering
\caption{Dataset Statistics.}
\vspace{0.1cm}
\begin{tabular}{c|ccc}
\hline
\hline
\textbf{Datasets} & \textbf{Proteins} & \textbf{IMDB-B} & \textbf{IMDB-M} \\ \hline
\#Nodes (max)           & 620          & 136              & 89              \\
\#Nodes (avg)            & 39.06          & 19.77              & 13.00             \\
\#Graphs        & 1113          & 1000              & 1500              \\
\#Classes        & 2          & 2              & 3              \\\hline
\end{tabular}
\label{tab:cladataset}
\end{table}
\textbf{Datasets.} We use three popular graph classification datasets, including one bioinformatics dataset Proteins, and two social network datasets IMDB-Binary and IMDB-Multi \cite{yanardag2015deep} for evaluation. The details are shown in Table \ref{tab:cladataset}.

\noindent\textbf{Settings.} We use 80\% samples for training, 10\% samples for validating and the rest 10\% for testing. The baseline model is a two-layer GCN with 16 hidden dimension and a global mean pooling layer after the second graph convolution layer. A linear read-out layer is attached to the output of the GCN to generate predictions. We apply the same training settings for GCN and GCN+WT-AWP. We train both models for 200 epoches with the Adam optimizer, learning rate 0.01 and weight decay 0.0005. The best model is selected with only the validation accuracy. For each of GCN and GCN+WT-AWP we take 10 random initialization and report the average accuracy and standard deviation. The hyperparameters $(\lambda, \rho)$ of WT-AWP is (0.3, 0.5) for Proteins, (0.05, 0.1) for IMDB-M and (0.5, 0.1) for IMDB-B.

\end{document}